\documentclass{article}
 \usepackage[preprint]{neurips_2026}

\usepackage[utf8]{inputenc} 
\usepackage[T1]{fontenc}    
\usepackage{hyperref}       
\usepackage{url}            
\usepackage{booktabs}       
\usepackage{amsfonts}       
\usepackage{nicefrac}       
\usepackage{microtype}      
\usepackage{xcolor}         
\usepackage{graphicx}
\usepackage{orcidlink}
\usepackage{amsmath}
\usepackage{booktabs}
\usepackage{multirow}
\usepackage{soul}
\usepackage{colortbl}
\usepackage{caption}
\usepackage{pifont}
\usepackage{makecell}
\usepackage{colortbl}

\usepackage{titletoc}
\usepackage{titlesec}
\usepackage{threeparttable}

\usepackage{booktabs}
\usepackage{longtable}
\usepackage{array}
\usepackage{multirow, makecell}
\usepackage{colortbl}

\usepackage{enumitem}
\usepackage{pifont}
\usepackage{amssymb,amsthm}
\usepackage{mathtools}
\usepackage{bm}
\usepackage{wrapfig}
\usepackage{adjustbox}
\usepackage{pifont}
\definecolor{grayrow}{gray}{0.92}
\definecolor{fullrowgray}{RGB}{235,235,235}
\definecolor{bestred}{RGB}{180,0,0}
\definecolor{secondblue}{RGB}{0,70,160}

\definecolor{gradleft}{HTML}{8E82D9} 
\definecolor{gradright}{HTML}{3FE4CA} 

\newcommand{\gradientOneWorld}{%
  \textcolor{gradleft!100}{O}%
  \textcolor{gradleft!95!gradright}{n}%
  \textcolor{gradleft!90!gradright}{e} %
  \textcolor{gradleft!86!gradright}{W}%
  \textcolor{gradleft!87!gradright}{o}%
  \textcolor{gradleft!76!gradright}{r}%
  \textcolor{gradleft!71!gradright}{l}%
  \textcolor{gradleft!67!gradright}{d}%
  \textcolor{gradleft!62!gradright}{,} %
  \textcolor{gradleft!57!gradright}{D}%
  \textcolor{gradleft!52!gradright}{u}%
  \textcolor{gradleft!48!gradright}{a}%
  \textcolor{gradleft!43!gradright}{l} %
  \textcolor{gradleft!38!gradright}{T}%
  \textcolor{gradleft!33!gradright}{i}%
  \textcolor{gradleft!29!gradright}{m}%
  \textcolor{gradleft!24!gradright}{e}%
  \textcolor{gradleft!19!gradright}{l}%
  \textcolor{gradleft!14!gradright}{i}%
  \textcolor{gradleft!10!gradright}{n}%
  \textcolor{gradleft!5!gradright}{e}%
}
\usepackage{xspace} 
\newcommand{\shadow}{\textsc{Dust}\xspace} 
\newcommand{\best}[1]{\textcolor{bestred}{{#1}}}
\newcommand{\second}[1]{\textcolor{secondblue}{{#1}}}
\definecolor{OrangeVar}{RGB}{230,120,20}
\definecolor{BlueVar}{RGB}{30,100,200}
\newcommand{\vo}[1]{\textcolor{OrangeVar}{#1}}
\newcommand{\vb}[1]{\textcolor{BlueVar}{#1}}

\newtheorem{theorem}{Theorem}

\newtheorem{definition}[theorem]{Definition}
\newtheorem{assumption}{Assumption}

\usepackage{xcolor}
\usepackage{soul}

\title{
  \begin{minipage}[c]{0.16\textwidth}
    \includegraphics[width=\linewidth]{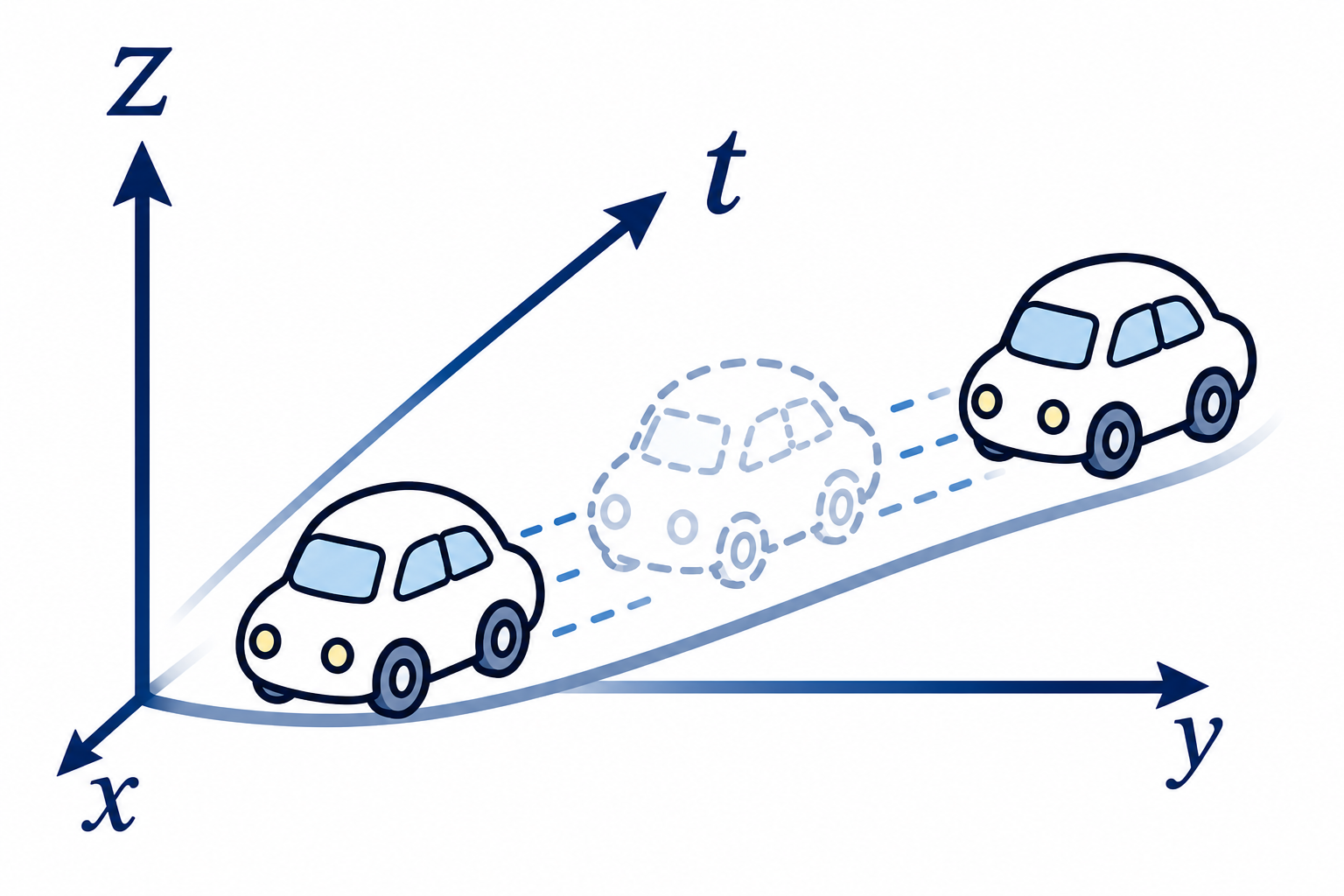}
  \end{minipage}
  \hspace{0.02\textwidth}
  \begin{minipage}[c]{0.8\textwidth}
\gradientOneWorld \textcolor{gradright}{:} Decoupled Spatio-Temporal Gaussian Scene Graph for 4D Cooperative Driving Reconstruction
  \end{minipage}
}
\author{%
Yulong Chen$^{\blacktriangle \spadesuit*}$ , 
Xiaoyun Dong$^{\spadesuit*}$, 
Haoyu Zhang$^{\spadesuit \clubsuit}$\thanks{Indicates equal contribution.}~~,
Zongxian Yang$^{\spadesuit}$, 
Lewei Xie$^{\blacktriangle}$, \\
\textbf{Xinke Li$^{\spadesuit}$}\thanks{Corresponding authors.}~~,
\textbf{Yifan Zhang$^{\blacktriangle}$}\footnotemark[2]~~,
\textbf{Kai Wang$^{\blacktriangle}$}\footnotemark[2]~~,
\textbf{Jianping Wang$^{\spadesuit}$} \\
  \normalfont{$^\blacktriangle$ City University of Hong Kong (Dongguan), Guangdong, China} \\
  \normalfont{$^\spadesuit$ City University of Hong Kong, Hong Kong, China} \\
  \normalfont{$^\clubsuit$ SLAI, Shenzhen, China} \\
{72405526@cityu-dg.edu.cn}\quad
{xiaodong8@cityu.edu.hk}\quad
{hzhang2838-c@my.cityu.edu.hk}\quad \\
{zongxian.yang@cityu-dg.edu.cn}\quad
{72404204@cityu-dg.edu.cn}\quad
{xinkeli@cityu.edu.hk}\quad\\
{kai.wang@cityu-dg.edu.cn}\quad
{yifan.zhang@cityu-dg.edu.cn}\quad
{jianwang@cityu.edu.hk}
}

\begin{document}
\maketitle
\begin{abstract}
Reconstructing dynamic scenes from Vehicle-to-Infrastructure Cooperative Autonomous Driving (VICAD) data is fundamentally complicated by temporal asynchrony: vehicle and infrastructure cameras operate on independent clocks, capturing the same dynamic agent such as cars and pedestrians at different physical times. Existing Gaussian Scene Graph methods implicitly assume synchronized observations and assign a single pose per agent per frame, which is an assumption that breaks in cooperative settings, where the resulting gradient conflicts cause severe ghosting on dynamic agents. We identify this as a \textit{representation-level failure}, not an optimization artifact: we prove that any single-timeline formulation incurs an irreducible photometric loss scaling quadratically with agent velocity and cross-source time offset. To resolve this, we propose \textbf{\textsc{Dust}} (\textbf{D}eco\textbf{U}pled \textbf{S}patio-\textbf{T}emporal) Gaussian Scene Graph for 4D Cooperative Driving Reconstruction. DUST Gaussian scene graph shares a canonical Gaussian set per agent for appearance consistency, while maintaining decouple pose trajectories aligned to each source's true captu
re timestamps. We prove that this decoupling enables the pose-gradient kernel block-diagonal, eliminating cross-source interference entirely. To make \textsc{Dust} practical, we further introduce a static anchor-based pose correction pipeline that corrects spatio misalignment between vehicle and infrastructure annotations, and a pose-regularized joint optimization scheme that prevents trajectory jitter and drift during early training. On 26 sequences from V2X-Seq,  \textsc{Dust} achieves state-of-the-art performance, improving dynamic-area PSNR by 3.2 dB over the strongest baseline and reducing Fréchet Video Distance by 37.7\%, with keeping robustness under larger temporal asynchrony.
\end{abstract}

\section{Introduction}
\label{sec:intro}
High-fidelity 4D scene reconstruction~\cite{chen2025omnire, kerbl20233d, yan2024street, zhou2024drivinggaussian} has become 
a practical alternative to real-world data collection for autonomous 
driving: it converts recorded driving logs into photorealistic digital 
environments that support novel-view synthesis (NVS), scene editing, 
and scalable simulation. A particularly compelling use case arises in 
Vehicle-to-Infrastructure Cooperative Autonomous Driving 
(VICAD)~\cite{lu2025vi, shi2022vips, xu2025instinct, yang2023bevheight, 
yu2024_univ2x}, where both a moving vehicle and fixed roadside cameras (i.e., infrastructure) 
observe the same scene. Reconstructing from both sources jointly 
enables observations that neither platform can produce alone, while also 
alleviating the shortage of real paired vehicle-infrastructure data, 
which is costly to collect at scale. 
\begin{figure}
    \centering
\includegraphics[width=\linewidth]{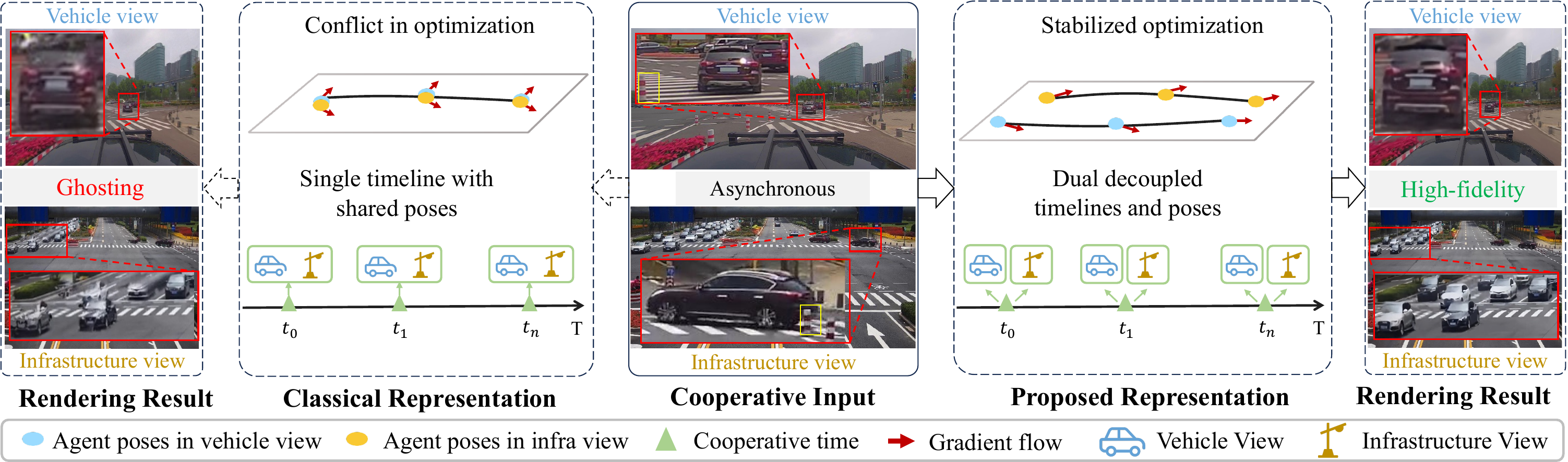}
     \caption{\textbf{Comparison of scene representations.}
     \textbf{Cooperative Input:} Vehicle and infrastructure sensors capture dynamic scenes asynchronously. \textbf{Top:} the black car has passed the pole; \textbf{Bottom}: it has not. \textbf{Classical Representation: } Aligning these asynchronous observations to one timeline creates conflicting gradients, leading to severe ghosting. \textbf{Proposed Representation:} Our DUST-GSG assigns separate pose timelines to each source while sharing canonical Gaussians, which removes gradient conflicts and enables stable, high-fidelity reconstruction.}
    \label{fig:mstgsg}
\vspace{-0.6cm}
\end{figure}

Recent driving reconstruction methods~\cite{chen2025omnire, yan2024street, zhou2024drivinggaussian} commonly adopt the Gaussian Scene Graph (GSG) to decompose a scene into a static background and dynamic agents, achieving strong results on single-platform data. These formulations all rest on an implicit assumption: at each frame, all observations share one timeline, and each dynamic agent is described by a single pose. This holds when sensors are driven by a common clock, but breaks down in cooperative settings, where vehicle and infrastructure cameras are triggered independently and their observations are inherently asynchronous. Forcing such asynchronous observations to share one pose induces conflicting gradients during optimization and produces ghosting on dynamic agents (Fig.~\ref{fig:mstgsg}). A concurrent effort on cooperative reconstruction~\cite{xu2025cruise} does not address this temporal asynchrony. The core challenge is therefore \textbf{\textit{how to faithfully represent asynchronous multi-source observations in a unified 4D scene}}.

\begin{figure}[h]
\vspace{-0.3cm}
    \centering
    \includegraphics[width=1\linewidth]{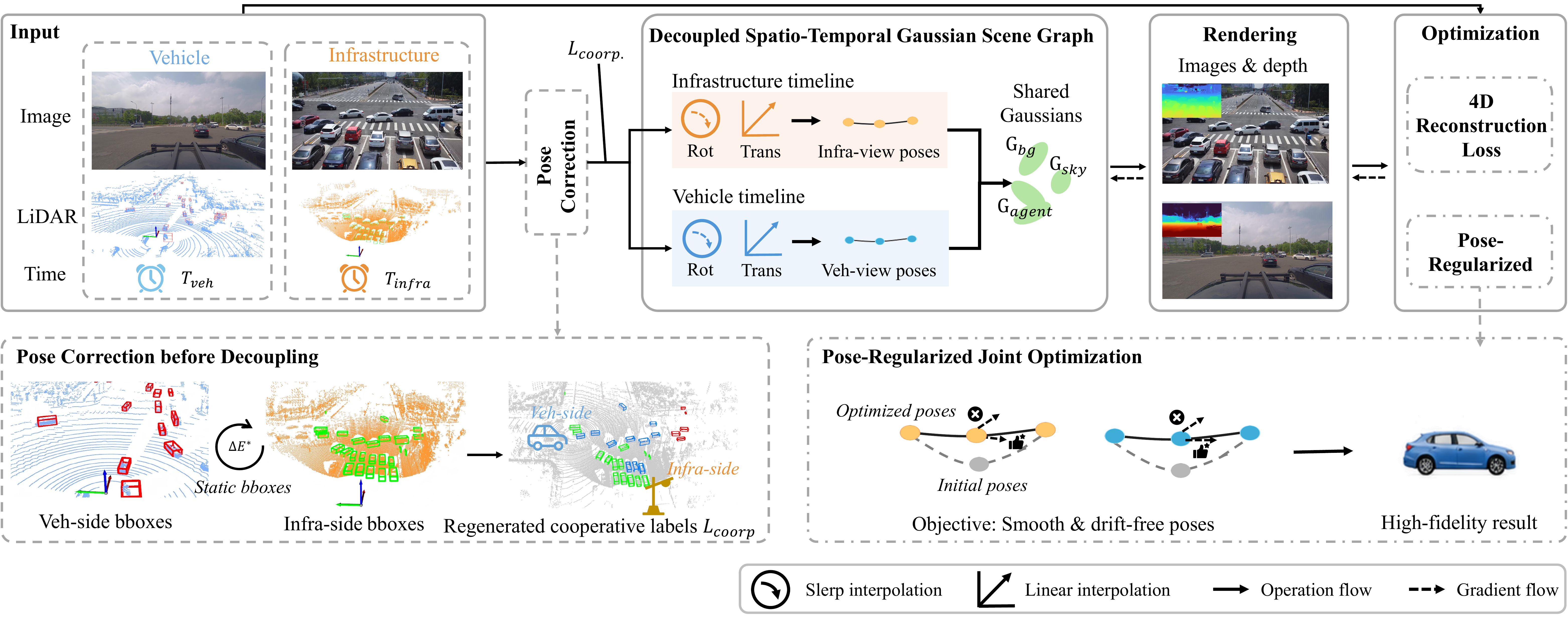}
    \caption{\textbf{Overview of \shadow.} \shadow first regenerates cooperative labels offline using co-visible static vehicles as anchors. The pose correction initialize a Decoupled Spatio-Temporal Gaussian Scene Graph, where each dynamic agent shares canonical Gaussians but maintains separate vehicle-side and infrastructure-side pose timelines. Joint optimization then refines Gaussian parameters and dual-timeline poses with pose regularization for stable cooperative reconstruction. }
    \label{fig:mts-gsg}
\vspace{-0.3cm}
\end{figure}

We propose \textbf{\shadow}, a 4D cooperative reconstruction framework based on a Decoupled Spatio-Temporal Gaussian Scene Graph (DUST-GSG). DUST-GSG models each dynamic agent with a shared set of canonical Gaussians to ensure appearance consistency, while assigning separate pose timelines to the vehicle and infrastructure sources (Fig.~\ref{fig:mts-gsg}). This decoupled design allows each source to render from its exact timestamp, directly eliminating the gradient conflicts caused by asynchronous observations. We further provide \textit{theoretical analysis} demonstrating that forcing a single timeline inherently causes reconstruction errors, whereas DUST-GSG mathematically resolves it. 
Beyond this theoretical design, applying our DUST-GSG to real cooperative data requires overcoming two problems. First, raw cooperative labels often contain spatio misalignment and provide poor pose initialization for DUST-GSG. We resolve this through a pose correction method. We use static vehicles as anchors to align labels from two sources, then regenerate cooperative labels for accurate initialization. Second, during reconstruction training, optimizing poses from pure image supervision easily leads to temporal drift and jitter. We address this by introducing a joint optimization scheme with pose regularization. This scheme enforces smooth motion across frames and prevents the poses from drifting away from their initial states. Combined with these two designs, \shadow ensures robust and high-fidelity cooperative reconstruction under asynchronous conditions.

Our main contributions are as follows:
\begin{itemize}[leftmargin=*,nosep]
\item We propose \shadow, a cooperative reconstruction paradigm using a Decoupled Spatio-Temporal Gaussian Scene Graph. We also provide a theoretical analysis of this paradigm.
\item We introduce a static anchor-based pose correction pipeline to refine vehicle-infrastructure alignment, subsequently regenerating cooperative labels offline to initialize agent poses.
\item We propose a pose-regularized joint optimization scheme for Gaussian parameters and agent poses, which enforces smooth 3D motion and effectively mitigates early-stage drift.
\item Evaluated on the V2X-Seq, \shadow achieves state-of-the-art performance in scene reconstruction and NVS, significantly outperforming previous methods in terms of dynamic area metrics (+\textbf{3.2 dB} PSNR and -\textbf{37.7\%} FVD) for reconstruction.
\end{itemize}

\section{Related Work}
\noindent\textbf{\textit{4D Driving Scene Reconstruction.}}
Recent driving reconstruction methods~\cite{cao2023hexplane, fridovich2023k, gao2021dynamic, kerbl20233d, li2021neural, pumarola2021d, wu20244d} build upon NeRF and 3D Gaussian Splatting to model dynamic urban environments. A dominant paradigm decomposes the scene into static backgrounds and dynamic agents. OMNIRE\cite{chen2025omnire}, StreetGS\cite{yan2024street}, and DrivingGaussian\cite{zhou2024drivinggaussian} track vehicles using explicit 3D tracking boxes and render them from canonical spaces. Other approaches like PVG\cite{chen2026periodic} avoid tracking boxes to model general scene vibrations. 
Extending these representations to cooperative reconstruction is an emerging frontier. For example, CRUISE~\cite{xu2025cruise} explores generative scene editing for cooperative scenarios. However, all these existing frameworks rely on a single timeline constraint. They assume perfectly synchronized cameras across all observations.
In real cooperative driving systems, vehicle and infrastructure sensors operate on independent hardware clocks. This hardware independence creates natural time gaps between cross observations. When existing methods force these asynchronous inputs into a single shared timeline, they inevitably produce severe artifacts. Our work directly targets this fundamental system limitation.

\noindent\textbf{\textit{Vehicle-Infrastructe Cooperative Autonomous Driving.}}
Cooperative systems share sensor data across vehicles and infrastructure to overcome physical line of sight limitations. Methods like DiscoNet~\cite{li2021learning} and V2X-ViT~\cite{xu2022v2x} exchange deep features to improve object detection. Other works optimize communication bandwidth~\cite{hu2022where2comm} or address spatial misalignment~\cite{lu2022robust}.
However, developing and evaluating these algorithms requires massive amounts of driving data. Most existing models are trained on purely synthetic datasets~\cite{xiang2024v2x, xu2022v2x}. These virtual environments fail to capture natural sensor data and complex physical constraints. 
While datasets like DAIR-V2X~\cite{dair-v2x} and V2X-Seq~\cite{v2x-seq} solve this by providing real world observations,  datasets cannot simulate new behaviors. Therefore, the cooperative autonomous driving industry urgently needs highly realistic simulation environments for algorithm development and testing. Our framework transforms real sequential data into fully controllable digital twins to directly serve this exact purpose.

\section{Method}

Our proposed \shadow framework uses inputs from both the vehicle and the infrastructure, including image sequences, camera poses, and cooperative tracking bounding boxes. Our goal is to learn a shared 4D Gaussian scene representation supervised by data from both sources. This directly supports novel view synthesis, scene editing, and driving simulation. First, Sec.\ref{sec:mtsgsg} introduces the Decoupled Spatio-Temporal Gaussian Scene Graph (DUST-GSG). Next, Sec.\ref{sec:theory} theoretically proves that DUST-GSG resolves the reconstruction errors caused by single-timeline modeling. Finally, to make DUST-GSG practical, Sec.\ref{sec:3_3} further introduces a pose correction step for accurate DUST-GSG initialization, and a joint optimization scheme that stabilizes the overall training process.
\subsection{Decoupled spatio-Temporal Gaussian Scene Graph} 
\label{sec:mtsgsg}
\textbf{\textit{Gaussian Scene Graph.}}
The Gaussian Scene Graph (GSG)~\cite{chen2025omnire,yan2024street,zhou2024drivinggaussian}
organizes a driving scene into a structured graph of 3D Gaussian nodes,
enabling joint reconstruction of static environments and dynamic agents.
It decomposes the scene into a time-invariant background node $\mathcal{V}_{\mathrm{bg}}$,
a sky node $\mathcal{V}_{\mathrm{sky}}$, and a set of dynamic agent nodes
$\{\mathcal{V}_{\mathrm{ag}}\}$ for moving objects such as vehicles and pedestrians,
where each node is represented as a set of 3D Gaussian primitives $\mathcal{G}$.
The background and sky nodes are static and time-invariant. 
The dynamic agents are further classified into rigid agents, such as cars and trucks, and non-rigid agents, like pedestrians and cyclists. 
For these agents,the key insight is to represent each agent by a \emph{canonical}
Gaussian set $G_{ag}=\{\mathcal{G}_n\}_{n=1}^{N}$, a fixed appearance defined
in the agent's local frame, decoupled from a pose trajectory $T(t)\in\mathrm{SE}(3)$
that places the agent in the world at time $t$:
\begin{equation}
\mathcal{V}_{\mathrm{ag}}(t) = T(t)\otimes f_{t}(G_{ag}).
\end{equation}
Here, $f_{t}:\mathbb{G}\rightarrow\mathbb{G}$ models the internal shape deformation of the agent, and $\mathbb{G}=\bigcup_{N\in\mathbb{N}}\{\{\mathcal{G}_n\}_{n=1}^{N}\}$ denotes the space of 3D Gaussian sets. For rigid agents, $f_t$ is simply the identity mapping. For non-rigid agents, $f_t$ serves as a time-dependent deformation field~\cite{yang2024deformable} that captures local movements like limb motion.
This factorization of a shared canonical shape optimized across all frames and a per-frame pose learned from image supervision makes GSG efficient and expressive for single-vehicle driving reconstruction. 

\textbf{\textit{Limitation in VICAD.}}
In VICAD, however, vehicle and infrastructure cameras are triggered asynchronously:
each source $c\in\{\mathrm{veh},\mathrm{infra}\}$ captures the same agent at a different
physical time $t_i^c \neq t_i$.
For static elements this discrepancy is negligible, but dynamic agents physically occupy
different world positions at $t_i^{\mathrm{veh}}$ and $t_i^{\mathrm{infra}}$.
Forcing both sources to share a single pose $T(t_i)$ thus attributes contradictory image
gradients to the same Gaussian primitives, causing the canonical shape to receive
conflicting supervision and the reconstructed geometry to be systematically misaligned.

\textbf{\textit{Decoupled spatio-Temporal Gaussian Scene Graph (DUST-GSG).}}
To address this, DUST-GSG keeps the canonical Gaussians $G_{ag}$ and deformation field $f_t$
\emph{shared} across both sources for spatio consistency,
but \emph{splits} the pose trajectory into two source-specific sequences
$\{T^{\mathrm{veh}}(t_i^{\mathrm{veh}})\}$ and $\{T^{\mathrm{infra}}(t_i^{\mathrm{infra}})\}$
aligned to each source's true capture times. The pose trajectory is initialized from cooperative labels. 
These labels include all visible agents from both observations. For each agent, the labels provide a sequence of  3D bounding boxes that track its position and orientation over time. To obtain the initial pose $T^c(t_i^c)$ for a specific source $c$ at query time $t_i^c$, we interpolate between these bounding boxes. We compute the position using linear interpolation and the rotation using spherical linear interpolation ($\operatorname{Slerp}$~\cite{shoemake1985animating}).
Each dynamic agent node is defined as a triplet:
\begin{equation}
\mathcal{V}_{\mathrm{ag}}^{\mathrm{DST}} :=
\Bigl\{\, G_{ag},\;\;
\{T^{\mathrm{veh}}(t_i^{\mathrm{veh}})\}_{i=0}^{K-1},\;\;
\{T^{\mathrm{infra}}(t_i^{\mathrm{infra}})\}_{i=0}^{K-1} \,\Bigr\},
\end{equation}

At render time, source $c$ queries its own trajectory at its own capture time $t_i^c$:
\begin{equation}
\mathcal{V}_{\mathrm{ag}}^{c}(t_i^c) = T^c(t_i^c) \otimes f_{t_i^c}(G_{ag}),
\qquad c \in \{\mathrm{veh},\mathrm{infra}\}.
\end{equation}
This factorization of a shared canonical shape and source-specific pose trajectories
forms the foundation of our DUST-GSG.
Following OMNIRE~\cite{chen2025omnire}, we optimize all Gaussian parameters
via differentiable rendering, but extend the optimization to jointly supervise
the shared canonical Gaussians $G_{ag}$ using image observations from
both the vehicle and infrastructure sources.

\subsection{Theoretical Justification of DUST-GSG}
\label{sec:theory}

Single-timeline methods fail in VICAD reconstruction because asynchronous sources
observe the same agent at different physical times.
Let $\Delta\tau = t_i^{\mathrm{veh}} - t_i^{\mathrm{infra}}$ be the
capture-time offset at cooperative timestamp $t_i$.
We formalize this failure and its remedy below;
full proofs are in Appendix~\ref{app:proof_gradient}.

\begin{theorem}\label{thm:gradient_decoupling}
Consider a dynamic agent with canonical Gaussians
$G_{ag}=\{\mathcal{G}_n\}_{n=1}^{N}$ (Eq.~(1)) moving at
velocity $\boldsymbol{v}\in\mathbb{R}^{3}$ during offset $\Delta\tau$.
Let $\mathcal{L}$ be the total photometric reconstruction loss
over both sources.
The following properties hold under local linearized rendering.

\textbf{1. Irreducible Error of Single Timeline.}
If both sources share a single pose $T(t_i)\in \text{SE}(3)$,
the optimal loss satisfies
\begin{equation}
    \mathcal{L}_{\mathrm{single}}^{*}
    \;\ge\;
    \frac{|\Delta\tau|^{2}\|\boldsymbol{v}\|^{2}}{4}
    \sum_{n=1}^{N}\lambda_{n}
    \;>\;0,
\end{equation}
where $\lambda_{n}>0$ is the minimum Fisher-information eigenvalue
of $\mathcal{G}_n$, quantifying its photometric sensitivity to displacement.

\textbf{2. Decoupling Property of DUST-GSG.}
Under $\mathcal{V}_{\mathrm{ag}}^{\mathrm{DST}}$ with source-specific
trajectories $\{T^{\mathrm{veh}}(t_i^{\mathrm{veh}})\}$ and
$\{T^{\mathrm{infra}}(t_i^{\mathrm{infra}})\}$,
a zero-loss solution exists ($\mathcal{L}_{\mathrm{DST}}^{*}=0$),
and the two pose sequences optimize independently:
a gradient update to $\{T^{\mathrm{veh}}\}$ induces no change
in the infrastructure loss, and vice versa.
\end{theorem}

\textbf{\textit{Discussion.}}
\ding{182} Property~1 shows the failure is \textbf{\textit{representational}}: no choice
of $G_{ag}$ can drive the loss to zero, and the irreducible error
grows quadratically with both $\|\boldsymbol{v}\|$ and $|\Delta\tau|$.
For a vehicle at 10\,m/s with $|\Delta\tau|=70$\,ms,
the implied misalignment reaches 0.7\,m, well beyond the support
radius of a single Gaussian, which directly causes the ghosting
in Fig.~\ref{fig:mstgsg}.
\ding{183} Property~2 shows DUST-GSG eliminates this at the representation level:
source-specific poses absorb the temporal gap, so $G_{ag}$ encodes
only intrinsic geometry and both timelines converge without mutual interference.
This motivates the pose correction and regularization strategies
described in the following.

\subsection{More Techniques for DUST-GSG} 
\label{sec:3_3}
\textbf{\textit{Pose Correction before Decoupling.}} Cooperative labels, i.e., bounding boxes, 
in dataset~\cite{dair-v2x} merge vehicle and infrastructure annotations based on provided 
infrastructure-to-vehicle poses. However, small pose errors often cause noticeable misalignment 
between the two sources. Because DUST-GSG uses these labels to initialize the pose trajectories 
$T^c(t_i^c)$, inaccurate labels will degrade reconstruction quality. To address this, we 
introduce an offline pose correction step before decoupling.

At each cooperative timestamp $t_i$, let $E_i\in \text{SE}(3)$ be the infrastructure-to-vehicle 
pose. We use co-visible static vehicles as geometric anchors to avoid dynamic interference. We 
match these static vehicles across both views using the Hungarian method~\cite{Kuhn}, and let 
$\mathcal{M}_i$ denote the set of matched vehicles at timestamp $t_i$. For a matched vehicle 
$m \in \mathcal{M}_i$, let $\mathbf{X}^{\mathrm{veh}}_{m}(t_i), 
\mathbf{X}^{\mathrm{infra}}_{m}(t_i) \in \mathbb{R}^{8\times 3}$ be the 3D bounding box corner 
matrices in the vehicle and infrastructure coordinates, respectively. We estimate a 6-DoF~\cite{zhou2019continuity} 
pose correction $\Delta E_i\in \text{SE}(3)$ by minimizing the corner alignment error:
\begin{equation}
\Delta E_i^*
=
\arg\min_{\Delta E_i\in \text{SE}(3)}
\frac{1}{|\mathcal{M}_i|}
\sum_{m \in \mathcal{M}_i}
\left\|
\mathbf{X}^{\mathrm{veh}}_{m}(t_i)
-
\Delta E_i E_i \mathbf{X}^{\mathrm{infra}}_{m}(t_i)
\right\|_F^2.
\end{equation}

We optimize this using L-BFGS~\cite{saputro2017limited}. Multiplying the optimal correction $\Delta E_i^*$ by the original pose $E_i$ gives the refined the pose. For co-visible objects, we use this refined pose to project infrastructure annotations into the vehicle coordinate system to form cooperative annotation. Objects observed by only one source are kept after being transformed into the unified coordinate system. 

Based on these regenerated labels, we interpolate poses to fill short missing segments in agent tracks. For an agent $a$ missing a pose at time $t_i$, bounded by a preceding valid timestamp $t_s$ and a subsequent valid timestamp $t_e$, we estimate the intermediate pose $T_a(t_i)$. We compute the translation via linear interpolation and the rotation via $\operatorname{Slerp}$~\cite{shoemake1985animating} between these two bounding poses. We only fill gaps of up to two frames. This procedure produces the pose correction for accurate DUST-GSG initialization.

\textbf{\textit{Pose-Regularized Joint Optimization.}} 
DUST-GSG shares the canonical Gaussians of each dynamic agent across views while maintaining source-specific pose trajectories in dual timelines. Although pose correction provide a good initialization, directly optimizing poses using only image supervision causes drift and frame-to-frame jitter. We therefore introduce a pose regularization loss. Let $T_a^c(t_i^c)$ be the initialized pose of agent $a$ in source $c \in \{\mathrm{veh}, \mathrm{infra}\}$, and $\tilde T_a^c(t_i^c)$ be its optimized pose. 

First, we force the optimized 3D trajectory to be smooth over time. Specifically, an agent's position at a given frame should closely match the linear interpolation of its past and future positions. For source $c$ and timestamp $t_i^c$, let $\mathcal A_i^c$ be the set of valid dynamic agents. We define the smoothness loss as:
\begin{equation}
\mathcal L_{\mathrm{smooth}}
=
\frac{1}{|\mathcal A_i^c|}
\sum_{a\in\mathcal A_i^c}
\left\|
\tilde T_a^c(t_i^c)
-
\left[
\tilde T_a^c(t_{i-1}^c)
+
w
\left(
\tilde T_a^c(t_{i+1}^c)
-
\tilde T_a^c(t_{i-1}^c)
\right)
\right]
\right\|_2,
\end{equation}
where $w = \frac{t_i^c-t_{i-1}^c}{t_{i+1}^c-t_{i-1}^c}$. Agents lacking valid past or future poses are excluded from $\mathcal A_i^c$ for this calculation. This term is applied independently to both timelines.

Second, we constrain the optimized 3D positions to remain close to their initialized positions during early training:
\begin{equation}
\mathcal L_{\mathrm{drift}}
=
\gamma(s)
\frac{1}{|\mathcal A_i^c|}
\sum_{a\in\mathcal A_i^c}
\left\|
\tilde T_a^c(t_i^c)
-
T_a^c(t_i^c)
\right\|_2 ,
\end{equation}
where $s$ is the current training step. The weight $\gamma(s)$ linearly decays to zero during training. This prevents poses from deviating before the Gaussian attributes become stable. As training proceeds, the constraint vanishes and image supervision naturally refines the poses.

The final training objective combines image rendering, geometric supervision, standard regularization, and the proposed pose regularization:
\begin{equation}
\mathcal L
=
\mathcal L_{\mathrm{image}}
+
\lambda_{\mathrm{depth}}\mathcal L_{\mathrm{depth}}
+
\lambda_{\mathrm{opacity}}\mathcal L_{\mathrm{opacity}}
+
\mathcal L_{\mathrm{reg}}
+
\lambda_{\mathrm{smooth}}\mathcal L_{\mathrm{smooth}}
+
\lambda_{\mathrm{drift}}\mathcal L_{\mathrm{drift}} .
\end{equation}
Here, $\mathcal L_{\mathrm{image}}$ is the image reconstruction loss based on L1 and SSIM. $\mathcal L_{\mathrm{depth}}$ uses sparse LiDAR supervision from both vehicle and infrastructure sensors. $\mathcal L_{\mathrm{opacity}}$ regularizes the rendered opacity. $\mathcal L_{\mathrm{reg}}$ contains the standard Gaussian regularization terms~\cite{chen2025omnire}. Details are provided in the Appendix~\ref{app:implementation_details}.

\section{Experiement}
In this section, we benchmark the reconstruction capabilities of \shadow against prior methods, focusing on dynamic scene reconstruction and novel view synthesis. 

\subsection{Experimental Setups}
\label{exp:setting}
\noindent\textbf{\textit{Dataset.}}
We evaluate our method on 26 diverse sequences selected from the large-scale V2X-Seq dataset~\cite{v2x-seq}. These sequences cover various times of day, weather and traffic densities~\cite{yang2023emernerf,9812038} (see Appendix~\ref{app:eva} for the full list). All images are processed at their original resolution of $1080{\times}1920$. To ensure spatio consistency, we regenerate the cooperative annotations for all sequences.

\noindent\textbf{\textit{Evaluation.}} Each cooperative timestamp provides a paired vehicle and infrastructure image. For scene reconstruction, we use all frames for both training and evaluation. For novel view synthesis, we hold out every 10th timestamp for testing, excluding both the vehicle and infrastructure images at these specific timestamps from the training set.
We measure reconstruction quality using PSNR, SSIM, and LPIPS. Since dynamic objects represent the core challenge in asynchronous cooperative reconstruction, we specifically report these metrics on dynamic areas, masked using the provided 2D bounding boxes~\cite{v2x-seq}. Furthermore, we report FVD~\cite{unterthiner2018towards} and RAFT-EPE~\cite{teed2020raft} on these dynamic areas to evaluate temporal and motion consistency. 

\noindent\textbf{\textit{Baselines.}}
We compare \shadow against several Gaussian Splatting approaches: PVG~\cite{chen2026periodic}, OMNIRE~\cite{chen2025omnire}, 3DGS~\cite{kerbl20233d}, CRUISE~\cite{xu2025cruise}, StreetGS~\cite{yan2024street}, and DeformableGS~\cite{yang2024deformable}. Among methods compared, for PVG, OMNIRE, StreetGS and DeformableGS, using their official implementations from the OMNIRE toolbox~\cite{chen2025omnire}. For a strictly fair comparison, all baselines are trained from scratch using the same data splits, evaluation protocols, and our regenerated annotations.

\noindent\textbf{\textit{Implementation Details.}}
We train all methods for $30{,}000$ iterations per scene on a single NVIDIA A100 GPU. For \shadow, the regularization coefficents are $\lambda_{\text{smooth}}=0.01$ and $\lambda_{\text{drift}}=0.01$. During inference, our model achieves real-time rendering at ${\sim}20$ FPS at $1080{\times}1920$ resolution. 

Details on the datasets, metrics, baselines, and implementation can be found in Appendix~\ref{app:implementation_details} and~\ref{app:eva}.
\subsection{Main Results}

\begin{table}[htbp]
\centering
\caption{\textbf{Performance comparison of scene reconstruction and NVS on the V2X-Seq dataset.} \shadow outperforms prior Gaussian methods under all conditions, achieving substantial improvement in dynamic areas. Cls and Cond denote sequence categories and conditions. Full averages all 26 sequences. \textcolor{bestred}{\textbf{Red}} and \textcolor{secondblue}{\textbf{Blue}} indicate the best and second-best results in a Cond line or Full line.}
\renewcommand{\arraystretch}{1.1}
\setlength{\tabcolsep}{2.5pt}
\resizebox{\textwidth}{!}{%
\begin{tabular}{l c c |ccc|cccc|ccc|cccc}

\toprule[1.2pt]

\multirow{3}{*}{\textbf{Method}}
& \multirow{3}{*}{\textbf{Cls.}}
& \multirow{3}{*}{\textbf{Cond.}}
& \multicolumn{7}{c|}{\textbf{Scene Reconstruction}}
& \multicolumn{7}{c}{\textbf{Novel View Synthesis}} \\

\cmidrule(lr){4-10} \cmidrule(lr){11-17}

& &
& \multicolumn{3}{c|}{\textit{Full Image}}
& \multicolumn{4}{c|}{\textit{Dynamic Area}}
& \multicolumn{3}{c|}{\textit{Full Image}}
& \multicolumn{4}{c}{\textit{Dynamic Area}} \\

\cmidrule(lr){4-6}  \cmidrule(lr){7-10}
\cmidrule(lr){11-13} \cmidrule(lr){14-17}

& &
& {\scriptsize PSNR$\uparrow$} & {\scriptsize SSIM$\uparrow$} & {\scriptsize LPIPS$\downarrow$} 
& {\scriptsize PSNR$\uparrow$} & {\scriptsize SSIM$\uparrow$} & {\scriptsize FVD$\downarrow$}  & {\scriptsize EPE$\downarrow$}
& {\scriptsize PSNR$\uparrow$} & {\scriptsize SSIM$\uparrow$} & {\scriptsize LPIPS$\downarrow$} 
& {\scriptsize PSNR$\uparrow$} & {\scriptsize SSIM$\uparrow$} & {\scriptsize FVD$\downarrow$}  & {\scriptsize EPE$\downarrow$} \\

\midrule[1.2pt]

\multirow{7}{*}{\makecell{\quad 3DGS~\cite{kerbl20233d} \\ {(2023)}}}
& \multirow{2}{*}{Time} & Noon
& 21.81 & 0.784 & 0.318 & 14.88 & 0.376 & 555.0 & 26.92
& 18.14 & 0.736 & 0.358 & 13.51 & 0.362 & 613.3 & 56.56 \\
&& Dusk
& 22.10 & 0.809 & 0.310 & 14.03 & 0.336 & 465.3 & 25.84
& 17.92 & 0.755 & 0.358 & 12.30 & 0.311 & 1110.6 & 63.48 \\
\cmidrule(lr){2-17}
& \multirow{2}{*}{Weather} & Normal
& 22.57 & 0.805 & 0.313 & 16.67 & 0.449 & 345.3 & 16.90
& 19.20 & 0.754 & 0.342 & 15.56 & 0.436 & 475.9 & 50.47 \\
&& Rainy
& 23.98 & 0.853 & 0.285 & 17.20 & 0.512 & 231.6 & 20.17
& 18.62 & 0.786 & 0.327 & 14.95 & 0.498 & 612.2 & 62.22 \\
\cmidrule(lr){2-17}
& \multirow{2}{*}{Crowd} & Low
& 26.03 & 0.842 & 0.280 & 17.02 & 0.465 & 225.6 & 24.65
& 22.77 & 0.803 & 0.309 & 15.73 & 0.459 & 380.3 & 71.86 \\
&& High
& 20.92 & 0.784 & 0.328 & 14.98 & 0.397 & 794.6 & 27.28
& 18.20 & 0.743 & 0.357 & 13.95 & 0.384 & 893.1 & 59.59 \\
\cmidrule(lr){2-17}
& \multicolumn{2}{c|}{\cellcolor{fullrowgray}\textit{Full}}
& \cellcolor{fullrowgray}22.72 & \cellcolor{fullrowgray}0.812 & \cellcolor{fullrowgray}0.307 
& \cellcolor{fullrowgray}15.82 & \cellcolor{fullrowgray}0.425 & \cellcolor{fullrowgray}433.2 & \cellcolor{fullrowgray}22.62
& \cellcolor{fullrowgray}18.91 & \cellcolor{fullrowgray}0.761 & \cellcolor{fullrowgray} 0.343
& \cellcolor{fullrowgray}14.32 & \cellcolor{fullrowgray}0.409 & \cellcolor{fullrowgray}682.2 & \cellcolor{fullrowgray}58.52 \\

\midrule

\multirow{7}{*}{\makecell{DeformGS~\cite{yang2024deformable} \\ {(2024)}}}
& \multirow{2}{*}{Time} & Noon
& 26.90 & 0.860 & 0.141 & 18.29 & 0.563 & 501.2 & 3.61
& 24.08 & 0.768 & 0.166 & 16.76 & 0.418 & 536.3 & 15.95 \\
&& Dusk
& 26.35 & 0.875 & 0.128 & 16.24 & 0.427 & 445.1 & 5.54
& 24.80 & 0.823 & 0.140 & 15.56 & 0.346 & 931.1 & 19.41 \\
\cmidrule(lr){2-17}
& \multirow{2}{*}{Weather} & Normal
& 27.83 & 0.884 & 0.121 & 19.68 & 0.625 & 294.6 & 3.55
& 25.09 & 0.803 & 0.145 & 18.43 & 0.501 & 419.9 & 15.33 \\
&& Rainy
& 29.22 & 0.899 & 0.131 & 20.36 & 0.640 & 181.2 & 3.81
& 26.22 & 0.835 & 0.150 & 18.56 & 0.511 & 469.7 & 16.70 \\
\cmidrule(lr){2-17}
& \multirow{2}{*}{Crowd} & Low
& 30.51 & \best{0.908} & \second{0.099} & 20.13 & 0.644 & 194.2 & 3.04
& 27.09 & \second{0.831} & \second{0.120} & 18.50 & 0.497 & 369.3 & 14.88 \\
&& High
& 25.28 & 0.841 & 0.174 & 17.09 & 0.504 & 608.5 & 5.45
& 23.38 & 0.771 & 0.182 & 16.22 & 0.397 & 724.6 & 19.53 \\
\cmidrule(lr){2-17}
& \multicolumn{2}{c|}{\cellcolor{fullrowgray}\textit{Full}}
& \cellcolor{fullrowgray}27.61 & \cellcolor{fullrowgray}0.877 & \cellcolor{fullrowgray}0.135 
& \cellcolor{fullrowgray}18.74 & \cellcolor{fullrowgray}0.572 & \cellcolor{fullrowgray}360.9 & \cellcolor{fullrowgray}4.3
& \cellcolor{fullrowgray}25.04 & \cellcolor{fullrowgray}0.804 & \cellcolor{fullrowgray}0.152 
& \cellcolor{fullrowgray}17.46 & \cellcolor{fullrowgray}0.450 & \cellcolor{fullrowgray}564.2 & \cellcolor{fullrowgray}16.84 \\

\midrule

\multirow{7}{*}{\makecell{StreetGS~\cite{yan2024street} \\ {(2024)}}}
& \multirow{2}{*}{Time} & Noon
& 27.96 & 0.870 & \second{0.124} & 20.81 & 0.729 & 172.9 & 1.97
& 25.21 & 0.781 & 0.146 & 18.40 & 0.518 & 210.1 & 12.29 \\
&& Dusk
& 28.38 & \second{0.889} & 0.106 & 20.84 & 0.715 & 102.6 & 2.41
& 26.48 & 0.832 & \second{0.121} & 19.29 & 0.589 & 234.8 & 13.80 \\
\cmidrule(lr){2-17}
& \multirow{2}{*}{Weather} & Normal
& 28.90 & \second{0.890} & \second{0.108} & 23.93 & 0.788 & 99.5 & 1.54
& 26.07 & 0.809 & 0.124 & 21.01 & \second{0.592} & 203.4 & 7.23 \\
&& Rainy
& 29.68 & 0.900 & 0.124 & 23.23 & 0.765 & 60.6 & 2.00
& 26.91 & 0.839 & \second{0.145} & 20.56 & 0.593 & 207.4 & 12.52 \\
\cmidrule(lr){2-17}
& \multirow{2}{*}{Crowd} & Low
& 30.89 & 0.903 & 0.102 & 23.14 & 0.771 & 46.5 & 2.11
& 27.71 & 0.828 & 0.121 & 20.28 & 0.561 & 123.7 & 12.21 \\
&& High
& 27.61 & 0.868 & 0.131 & 21.58 & 0.737 & 137.6 & 2.13
& 25.09 & 0.784 & 0.156 & 19.29 & 0.550 & 253.2 & 13.29 \\
\cmidrule(lr){2-17}
& \multicolumn{2}{c|}{\cellcolor{fullrowgray}\textit{Full}}
& \cellcolor{fullrowgray}28.88 & \cellcolor{fullrowgray}0.886 & \cellcolor{fullrowgray}0.116 
& \cellcolor{fullrowgray}22.54 & \cellcolor{fullrowgray}0.758 & \cellcolor{fullrowgray}103.0 & \cellcolor{fullrowgray}2.01
& \cellcolor{fullrowgray}26.20 & \cellcolor{fullrowgray}0.812 & \cellcolor{fullrowgray}0.135 
& \cellcolor{fullrowgray}20.01 & \cellcolor{fullrowgray}0.572 & \cellcolor{fullrowgray}214.8 & \cellcolor{fullrowgray}11.47 \\

\midrule

\multirow{7}{*}{\makecell{OMNIRE~\cite{chen2025omnire} \\ {(2025)}}}
& \multirow{2}{*}{Time} & Noon
& 28.16 & \second{0.871} & \best{0.122} & 20.97 & \second{0.736} & \second{158.7} & \second{1.85}
& 25.33 & \second{0.783} & \second{0.141} & 18.43 & 0.517 & \second{205.5} & \second{11.75} \\
&& Dusk
& \second{28.70} & \second{0.889} & \second{0.105} & \second{21.13} & \second{0.727} & \second{46.7} & \second{2.24}
& \second{26.74} & \second{0.833} & \second{0.121} & \second{19.53} & \second{0.599} & \second{71.4} & \second{13.22} \\
\cmidrule(lr){2-17}
& \multirow{2}{*}{Weather} & Normal
& \second{29.14} & 0.888 & \second{0.108} & \second{24.12} & \second{0.791} & \second{56.6} & \second{1.46}
& 26.20 & \second{0.810} & \second{0.122} & \second{21.02} & 0.590 & \second{112.7} & \second{7.05} \\
&& Rainy
& 30.06 & \second{0.902} & \second{0.122} & 24.04 & 0.779 & \second{35.1} & \second{1.79}
& 27.20 & 0.838 & 0.146 & 21.20 & 0.604 & \second{138.2} & \second{12.25} \\
\cmidrule(lr){2-17}
& \multirow{2}{*}{Crowd} & Low
& \second{31.27} & 0.904 & 0.101 & 23.95 & 0.792 & \second{25.8} & 2.02
& \second{27.95} & 0.828 & \second{0.120} & \second{20.94} & 0.585 & \second{78.5} & \second{11.76} \\
&& High
& \second{28.01} & \second{0.869} & \second{0.129} & \second{21.83} & \second{0.741} & \second{89.3} & \second{2.03}
& \second{25.34} & \second{0.788} & \second{0.146} & \second{19.42} & \second{0.551} & \second{170.5} & \second{12.58} \\
\cmidrule(lr){2-17}
& \multicolumn{2}{c|}{\cellcolor{fullrowgray}\textit{Full}}
& \cellcolor{fullrowgray}\second{29.20} & \cellcolor{fullrowgray}\second{0.887} & \cellcolor{fullrowgray}\second{0.114} 
& \cellcolor{fullrowgray}22.90 & \cellcolor{fullrowgray}\second{0.767} & \cellcolor{fullrowgray}\second{63.7} & \cellcolor{fullrowgray}\second{1.87}
& \cellcolor{fullrowgray}\second{26.41} & \cellcolor{fullrowgray}0.813 & \cellcolor{fullrowgray}\second{0.132} 
& \cellcolor{fullrowgray}\second{20.25} & \cellcolor{fullrowgray}\second{0.579} & \cellcolor{fullrowgray}\second{125.2} & \cellcolor{fullrowgray}\second{10.93} \\

\midrule

\multirow{7}{*}{\makecell{CRUISE~\cite{xu2025cruise} \\ {(2025)}}}
& \multirow{2}{*}{Time} & Noon
& 23.40 & 0.799 & 0.209 
& 17.15 & 0.490 & 378.1 & 5.44
& 22.60 & 0.765 & 0.220 
& 16.71 & 0.445 & 440.6 & 23.65 \\
&& Dusk
& 24.01 & 0.821 & 0.193 
& 18.91 & 0.591 & 147.9 & 5.45
& 23.51 & 0.804 & 0.196 
& 16.67 & 0.444 & 452.5 & 24.50 \\
\cmidrule(lr){2-17}
& \multirow{2}{*}{Weather} & Normal
& 24.19 & 0.822 & 0.184 
& 19.39 & 0.601 & 277.4 & 5.33
& 23.00 & 0.793 & 0.199 
& 18.39 & 0.514 & 336.7 & 25.08 \\
&& Rainy
& 25.48 & 0.856 & 0.176 
& 18.77 & 0.569 & 134.4 & 5.84
& 24.62 & 0.836 & 0.187 
& 18.27 & 0.518 & 313.5 & 24.81 \\
\cmidrule(lr){2-17}
& \multirow{2}{*}{Crowd} & Low
& 23.18 & 0.802 & 0.204 
& 17.84 & 0.513 & 134.8 & 4.95
& 25.75 & 0.831 & 0.168 
& 18.61 & 0.542 & 265.7 & 23.83 \\
&& High
& 26.98 & 0.861 & 0.149 
& 18.96 & 0.584 & 360.7 & 6.38
& 22.57 & 0.779 & 0.215 
& 17.24 & 0.466 & 440.7 & 26.48 \\
\cmidrule(lr){2-17}
& \multicolumn{2}{c|}{\cellcolor{fullrowgray}\textit{Full}}
& \cellcolor{fullrowgray}24.47 & \cellcolor{fullrowgray}0.826 & \cellcolor{fullrowgray}0.187 
& \cellcolor{fullrowgray}18.57 & \cellcolor{fullrowgray}0.561 & \cellcolor{fullrowgray}433.2 & \cellcolor{fullrowgray}22.62
& \cellcolor{fullrowgray}23.63 & \cellcolor{fullrowgray}0.801 & \cellcolor{fullrowgray}0.199 
& \cellcolor{fullrowgray}17.68 & \cellcolor{fullrowgray}0.488 & \cellcolor{fullrowgray}397.6 & \cellcolor{fullrowgray}24.49 \\

\midrule

\multirow{7}{*}{\makecell{\quad PVG~\cite{chen2026periodic} \\ {(2026)}}}
& \multirow{2}{*}{Time} & Noon
& \second{28.42} & 0.858 & 0.174 & \second{22.36} & 0.734 & 185.6 & 2.21
& \second{25.87} & \best{0.789} & 0.193 & \second{19.70} & \second{0.549} & 323.5 & 12.38 \\
&& Dusk
& 27.82 & 0.883 & 0.148 & 20.47 & 0.665 & 134.9 & 3.62
& 25.64 & 0.830 & 0.161 & 18.20 & 0.491 & 306.4 & 17.83 \\
\cmidrule(lr){2-17}
& \multirow{2}{*}{Weather} & Normal
& 29.01 & 0.880 & 0.150 & 23.74 & 0.772 & 137.3 & 2.23
& \second{26.28} & \best{0.817} & 0.159 & 20.90 & 0.591 & 226.5 & 12.36 \\
&& Rainy
& \best{30.46} & 0.900 & 0.149 & \second{24.61} & \second{0.789} & 51.5 & 2.39
& \second{27.37} & \best{0.848} & 0.164 & \second{21.38} & \second{0.609} & 194.6 & 13.75 \\
\cmidrule(lr){2-17}
& \multirow{2}{*}{Crowd} & Low
& \best{31.43} & 0.903 & 0.120 & \second{24.85} & \second{0.808} & 46.1 & \second{1.76}
& 27.53 & \best{0.835} & 0.137 & \second{21.12} & \second{0.602} & 121.4 & 14.57 \\
&& High
& 27.06 & 0.852 & 0.189 & 21.27 & 0.675 & 248.3 & 3.86
& 24.72 & 0.787 & 0.203 & 19.15 & 0.520 & 410.0 & 16.14 \\
\cmidrule(lr){2-17}
& \multicolumn{2}{c|}{\cellcolor{fullrowgray}\textit{Full}}
& \cellcolor{fullrowgray}28.97 & \cellcolor{fullrowgray}0.879 & \cellcolor{fullrowgray}0.157 
& \cellcolor{fullrowgray}\second{22.91} & \cellcolor{fullrowgray}0.742 & \cellcolor{fullrowgray}131.9 & \cellcolor{fullrowgray}2.77
& \cellcolor{fullrowgray}26.16 & \cellcolor{fullrowgray}\second{0.817} & \cellcolor{fullrowgray}0.171 
& \cellcolor{fullrowgray}20.09 & \cellcolor{fullrowgray}0.562 & \cellcolor{fullrowgray}262.3 & \cellcolor{fullrowgray}14.24 \\

\midrule

\multirow{7}{*}{\makecell{\quad \textbf{Ours} \\ {\quad (\textbf{\shadow})}}}
& \multirow{2}{*}{Time} & Noon
& \best{28.75} & \best{0.872} & \best{0.122} 
& \best{24.11} & \best{0.839} & \best{78.1} & \best{1.65}
& \best{25.94} & \best{0.789} & \best{0.137} 
& \best{20.04} & \best{0.584} & \best{167.5} & \best{9.14} \\
&& Dusk
& \best{29.05} & \best{0.892} & \best{0.102} 
& \best{24.57} & \best{0.844} & \best{27.8} & \best{2.04}
& \best{27.03} & \best{0.834} & \best{0.119} 
& \best{21.50} & \best{0.667} & \best{69.5} & \best{11.90} \\
\cmidrule(lr){2-17}
& \multirow{2}{*}{Weather} & Normal
& \best{29.42} & \best{0.891} & \best{0.105} 
& \best{26.85} & \best{0.871} & \best{45.1} & \best{1.31}
& \best{26.72} & \best{0.817} & \best{0.120} 
& \best{22.67} & \best{0.641} & \best{111.0} & \best{6.52} \\
&& Rainy
& \second{30.36} & \best{0.904} & \best{0.119} 
& \best{27.86} & \best{0.880} & \best{21.8} & \best{1.62}
& \best{27.49} & \second{0.842} & \best{0.141} 
& \best{22.44} & \best{0.640} & \best{84.6} & \best{11.82} \\
\cmidrule(lr){2-17}
& \multirow{2}{*}{Crowd} & Low
& \best{31.43} & \second{0.906} & \best{0.098} 
& \best{27.61} & \best{0.878} & \best{18.1} & \best{1.74}
& \best{28.18} & \second{0.831} & \best{0.118} 
& \best{22.40} & \best{0.627} & \best{46.0} & \best{11.44} \\
&& High
& \best{28.56} & \best{0.872} & \best{0.125} 
& \best{24.49} & \best{0.829} & \best{47.4} & \best{1.90}
& \best{25.90} & \best{0.794} & \best{0.145} 
& \best{21.01} & \best{0.613} & \best{105.3} & \best{10.80} \\
\cmidrule(lr){2-17}
& \multicolumn{2}{c|}{\cellcolor{fullrowgray}\textit{Full}}
& \cellcolor{fullrowgray}\best{29.55} & \cellcolor{fullrowgray}\best{0.889} & \cellcolor{fullrowgray}\best{0.112} 
& \cellcolor{fullrowgray}\best{26.11} & \cellcolor{fullrowgray}\best{0.860} & \cellcolor{fullrowgray}\best{39.7} & \cellcolor{fullrowgray}\best{1.70}
& \cellcolor{fullrowgray}\best{26.82} & \cellcolor{fullrowgray}\best{0.818} & \cellcolor{fullrowgray}\best{0.130} 
& \cellcolor{fullrowgray}\best{21.80} & \cellcolor{fullrowgray}\best{0.631} & \cellcolor{fullrowgray}\best{97.2} & \cellcolor{fullrowgray}\best{9.94} \\

\bottomrule[1.2pt]

\end{tabular}%
}
\label{tab1}
\end{table}

\begin{figure}
\vspace{-0.3cm}
\centering
\includegraphics[width=1\linewidth]{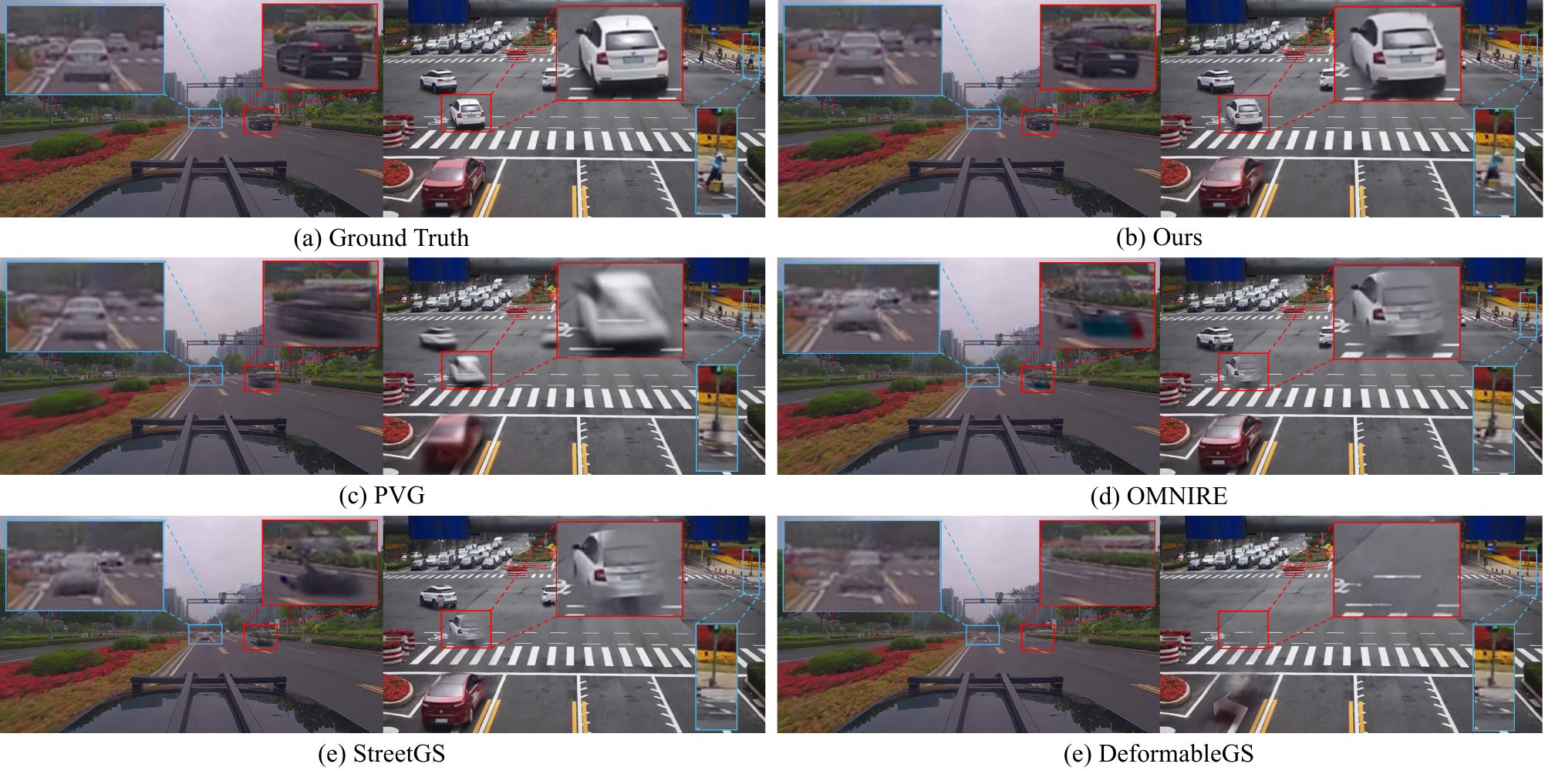}
\caption{\textbf{Qualitative comparison on dynamic agents under asynchronous observations.}
Compared to baselines, our method \shadow achieves less ghosting and more high-quality reconstruction of various agents, including vehicles and pedestrians. The insets highlight the details.}
\label{fig:fig3}
\vspace{-0.3cm}
\end{figure}

As shown in Tab.~\ref{tab1}, \shadow achieves state-of-the-art performance in dynamic areas and full images. For instance, our dynamic PSNR in scene reconstruction exceeds the second-best PVG by 3.2 dB. Previous methods like OMNIRE and StreetGS force asynchronous observations into a single timeline. This shared-pose assumption creates optimization conflicts for dynamic agents. By decoupling the timelines, our DUST-GSG directly resolves these conflicts and recovers accurate appearances.  
Multi-frame metrics further confirm this advantage. \shadow improves scene reconstruction FVD by 37.7\%, dropping it from 63.7 to 39.7. This temporal stability comes from both the DUST-GSG representation and our pose regularization, which jointly prevent trajectory jitter. Visual comparisons in Fig.~\ref{fig:fig3} align with these numbers. While baselines suffer from severe ghosting and blurring on moving vehicles and pedestrians, \shadow renders sharp and artifact-free dynamic agents.  

\subsection{Complex Driving Conditions}
Tab.~\ref{tab1} shows our method is robust in complex environments. The gap is clearest in crowded scenes, where dense traffic creates more asynchronous conflicts. Baselines struggle here, but \shadow achieves an FVD of 47.4, reducing OMNIRE's score of 89.3 by nearly 47\%. We also maintain a strong lead in dynamic areas during rainy weather and at dusk.

While \shadow ranks second in a few full-image metrics, the difference is marginal. For example, our full-image PSNR in rainy conditions is only 0.1 dB lower than PVG. This slight drop is expected because full-image scores are dominated by static backgrounds, which existing methods already model well. 

\subsection{Asynchronous Observations}
\begin{wrapfigure}{r}{0.55\textwidth}
\vspace{-1.4cm}
    \centering
    \includegraphics[width=\linewidth]{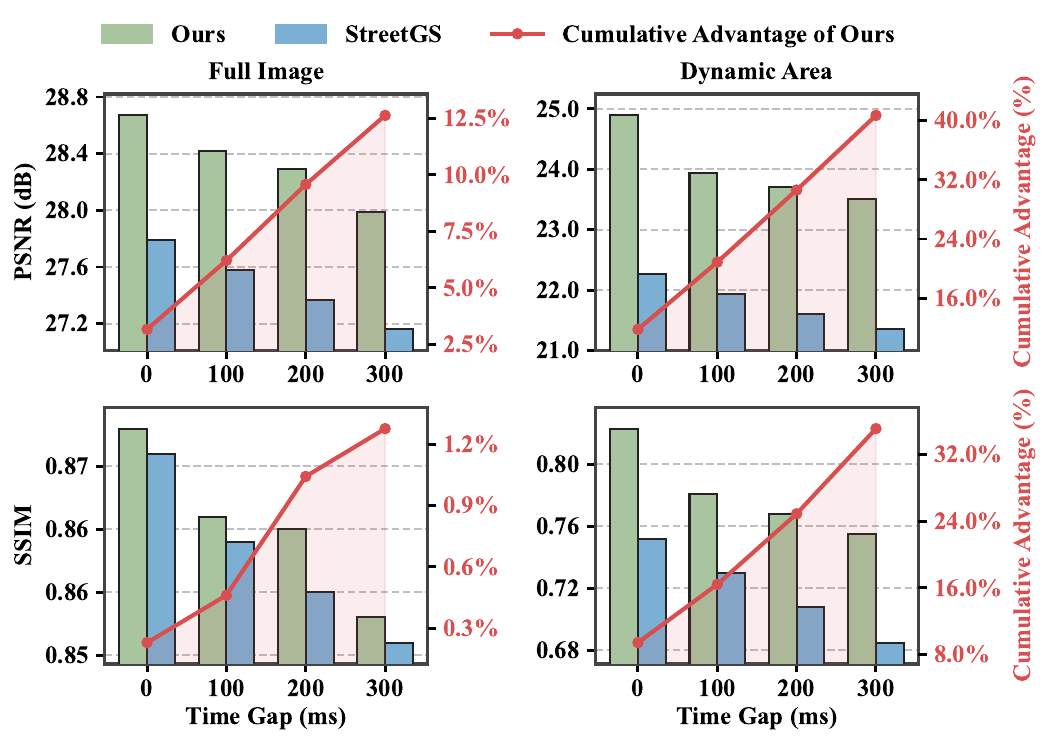}
    \caption{Performance comparison between Ours and StreetGS under increasing temporal asynchrony of 0-300 (Time Gap)\,ms. The cumulative growth rate of our method over StreetGS on the specified metrics is illustrated by the red curve.}
    \label{fig:fig5}
    \vspace{-0.6cm}
\end{wrapfigure}
To evaluate robustness against asynchronous data, we introduce
 time delays between the two observation sources. Specifically, we pair the vehicle frames with infrastructure frames shifted by one to three time steps. This setup creates physical time gaps ranging from zero to 300 milliseconds. Fig.~\ref{fig:fig5} compares \shadow and StreetGS using PSNR and SSIM for both full images and dynamic areas. As the time gap increases, StreetGS performance drops rapidly. In contrast, \shadow shows a much slower decay. This demonstrates that \shadow has stronger robustness to temporal asynchrony. The figure also plots our cumulative advantage. 
This metric grows steadily with larger time delays. The widening gap confirms that our decoupled design effectively handles severe asynchrony. It successfully prevents the motion artifacts common in single timeline methods.

\subsection{Ablation Studies}
\begin{figure}[h]
    \centering
    \includegraphics[width=0.8\linewidth]{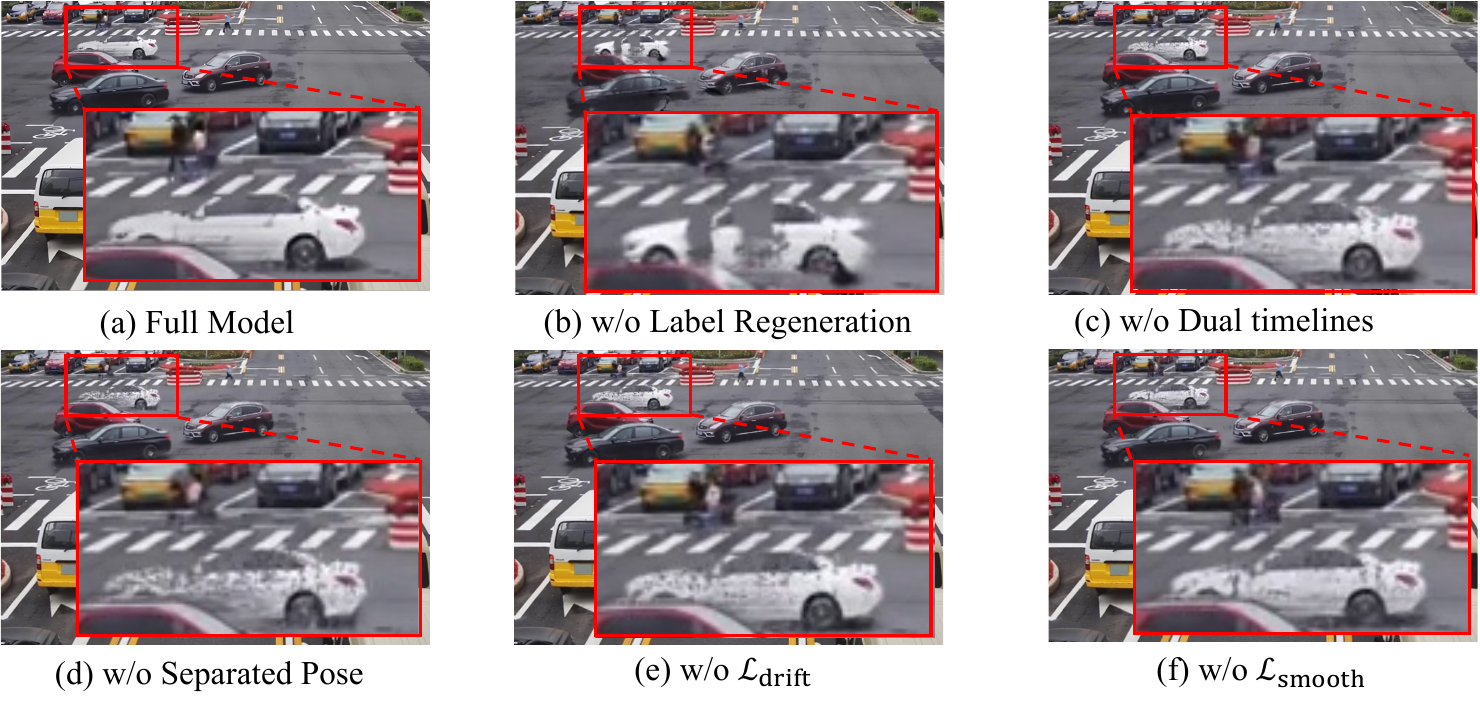}
    \caption{\textbf{Qualitative ablation study} results across infrastructure camera from the scene reconstruction task. (a) Our complete model. (b) Without pose initialization before decoupling. (c) Without dual timelines. (d) Without separated poses. (e) Without $\mathcal L_{\mathrm{drift}}$. (f) Without $\mathcal L_{\mathrm{smooth}}$.}
    \label{fig:fig4}
\vspace{-0.3cm}
\end{figure}

We conduct comprehensive ablation studies to evaluate the contribution of each module, with results shown in Tab.~\ref{tab2} and Fig.~\ref{fig:fig4}. The findings demonstrate the critical role of our proposed 
\begin{wraptable}{r}{0.55\textwidth}
\vspace{-4pt}
\footnotesize
\centering
\caption{Ablations on the High Crowd scenes subset. Dual Timelines and Separated are the part of DUST-GSG}
\begin{adjustbox}{width=0.52\textwidth, max width=0.52\textwidth}
\begin{tabular}{c c c c c c}
\toprule
\multirow{2}{*}{\textbf{Variant}} & \multicolumn{3}{c}{\textbf{Full Image}} & \multicolumn{2}{c}{\textbf{Dynamic Area}} \\
\cmidrule(lr){2-4} \cmidrule(lr){5-6}
                       & PSNR$\uparrow$ & SSIM$\uparrow$ & LPIPS$\downarrow$ & PSNR$\uparrow$ & SSIM$\uparrow$ \\
\midrule
Full Model             & \textbf{28.75} & \textbf{0.874} & \textbf{0.125}    & \textbf{25.12} & \textbf{0.837} \\
w/o Pose correction & 27.91          & 0.873          & 0.128             & 23.98          & 0.821          \\
w/o Dual Timelines     & 28.40          & 0.872          & 0.128             & 23.35          & 0.783          \\
w/o Separated Poses    & 28.19          & 0.870          & 0.131             & 22.34          & 0.742          \\
w/o $\mathcal L_{\mathrm{drift}}$   & 28.72 & 0.873 & 0.126 & 24.99 & 0.832 \\
w/o $\mathcal L_{\mathrm{smooth}}$  & 28.62 & 0.869 & 0.134 & 24.96 & 0.833 \\
\bottomrule
\end{tabular}
\end{adjustbox}
\vspace{-11pt}                         
\label{tab2}
\end{wraptable}
 DUST-GSG representation. Removing either the dual timelines or the separated poses causes severe performance drops, especially in dynamic areas where the vehicle PSNR falls from 25.12 to 23.35 and 22.34, respectively. This confirms that decoupling spatio and temporal states is essential for handling asynchronous cooperative data. 
Furthermore, the static anchor-based pose correction provides crucial initialization.  
Without it, the dynamic PSNR decreases by 1.14 dB. This proves that refining the raw dataset labels is a necessary step before initializing the Gaussian trajectories. Without the pose regularation ($\mathcal L_{\mathrm{smooth}}$), the LPIPS falls from 0.125 to 0.134. This demonstrates that applying the smooth regularization to the pose is crucial for refined reconstruction. Finally, while removing the pose regularization ($\mathcal L_{\mathrm{drift}}$) causes only a slight drop in PSNR, this term is vital for dynamic stability. As shown in Fig.~\ref{fig:fig4}e, without this constraint, unconstrained joint optimization leads to trajectory drift and jittering during early training. The regularization ensures the Gaussians learn correct appearances rather than compensating for incorrect positions. 
Together, these modules allow \shadow to effectively reconstruct dynamic objects and establish a strong decoupled paradigm for VICAD reconstruction.

\section{Conclusion}
We present \textbf{\shadow}. It is a cooperative reconstruction framework designed for asynchronous autonomous driving scenarios. We mathematically prove that using a single timeline inherently causes reconstruction errors. To resolve this, we introduce a decoupled scene graph. We combine this core design with anchor based pose correction and spatio pose regularization. Extensive experiments on the V2X Seq dataset show that our method achieves superior performance across various asynchronous conditions.
Our current framework has a  limitation. Our decoupled design successfully resolves the trajectory mismatch for dynamic agents. However, it does not explicitly handle the shape changes of deformable objects like pedestrians under asynchronous observations. Time delays can still cause appearance blur during rapid limb movements. Future research will explore feature based fine grained temporal deformation modeling.

\newpage
\medskip
\bibliographystyle{plain}
\bibliography{neurips_2026}

@inproceedings{chen2025omnire,
  title={Omnire: Omni urban scene reconstruction},
  author={Chen, Ziyu and Yang, Jiawei and Huang, Jiahui and De Lutio, Riccardo and Martinez Esturo, Janick and Ivanovic, Boris and Litany, Or and Gojcic, Zan and Fidler, Sanja and Pavone, Marco and others},
  booktitle={International Conference on Learning Representations},
  volume={2025},
  pages={85508--85527},
  year={2025}
}

@inproceedings{zhou2024drivinggaussian,
  title={Drivinggaussian: Composite gaussian splatting for surrounding dynamic autonomous driving scenes},
  author={Zhou, Xiaoyu and Lin, Zhiwei and Shan, Xiaojun and Wang, Yongtao and Sun, Deqing and Yang, Ming-Hsuan},
  booktitle={Proceedings of the IEEE/CVF Conference on Computer Vision and Pattern Recognition},
  pages={21634--21643},
  year={2024}
}

@inproceedings{zhou2019continuity,
  title={On the continuity of rotation representations in neural networks},
  author={Zhou, Yi and Barnes, Connelly and Lu, Jingwan and Yang, Jimei and Li, Hao},
  booktitle={Proceedings of the IEEE/CVF conference on computer vision and pattern recognition},
  pages={5745--5753},
  year={2019}
}

@inproceedings{saputro2017limited,
  title={Limited memory Broyden-Fletcher-Goldfarb-Shanno (L-BFGS) method for the parameter estimation on geographically weighted ordinal logistic regression model (GWOLR)},
  author={Saputro, Dewi Retno Sari and Widyaningsih, Purnami},
  booktitle={AIP conference proceedings},
  volume={1868},
  pages={040009},
  year={2017},
  organization={AIP Publishing LLC}
}

@inproceedings{shoemake1985animating,
  title={Animating rotation with quaternion curves},
  author={Shoemake, Ken},
  booktitle={Proceedings of the 12th annual conference on Computer graphics and interactive techniques},
  pages={245--254},
  year={1985}
}

@inproceedings{lu2025vi,
  title={VI-Planning: Infrastructure-Assisted Real-Time Planning Optimization for Autonomous Driving},
  author={Lu, Yang and Wang, Jie and Dong, Xiaoyun and Huang, Ziyao and Liu, Bingyi and Wu, Jen-Ming and Wang, Jianping},
  booktitle={Proceedings of the 31st Annual International Conference on Mobile Computing and Networking},
  pages={923--937},
  year={2025}
}

@inproceedings{yang2023bevheight,
    title={BEVHeight: A Robust Framework for Vision-based Roadside 3D Object Detection},
    author={Yang, Lei and Yu, Kaicheng and Tang, Tao and Li, Jun and Yuan, Kun and Wang, Li and Zhang, Xinyu and Chen, Peng},
    booktitle={IEEE/CVF Conf.~on Computer Vision and Pattern Recognition (CVPR)},
    month = mar,
    year={2023}
}

@inproceedings{yu2024_univ2x,
 title={End-to-End Autonomous Driving through V2X Cooperation}, 
 author={Haibao Yu and Wenxian Yang and Jiaru Zhong and Zhenwei Yang and Siqi Fan and Ping Luo and Zaiqing Nie},
 booktitle={The 39th Annual AAAI Conference on Artificial Intelligence},
 year={2025}
}

@inproceedings{shi2022vips,
  title={VIPS: Real-time perception fusion for infrastructure-assisted autonomous driving},
  author={Shi, Shuyao and Cui, Jiahe and Jiang, Zhehao and Yan, Zhenyu and Xing, Guoliang and Niu, Jianwei and Ouyang, Zhenchao},
  booktitle={Proceedings of the 28th annual international conference on mobile computing and networking},
  pages={133--146},
  year={2022}
}

@inproceedings{xu2025instinct,
  title={INSTINCT: Instance-Level Interaction Architecture for Query-Based Collaborative Perception},
  author={Xu, Yunjiang and Li, Lingzhi and Wang, Jin and Ouyang, Yupeng and Yang, Benyuan},
  booktitle={Proceedings of the IEEE/CVF International Conference on Computer Vision},
  pages={25464--25473},
  year={2025}
}

@inproceedings{dair-v2x,
  title={Dair-v2x: A large-scale dataset for vehicle-infrastructure cooperative 3d object detection},
  author={Yu, Haibao and Luo, Yizhen and Shu, Mao and Huo, Yiyi and Yang, Zebang and Shi, Yifeng and Guo, Zhenglong and Li, Hanyu and Hu, Xing and Yuan, Jirui and Nie, Zaiqing},
  booktitle={Proceedings of the IEEE/CVF Conference on Computer Vision and Pattern Recognition},
  pages={21361--21370},
  year={2022}
}

@inproceedings{v2x-seq,
  title={V2X-Seq: A large-scale sequential dataset for vehicle-infrastructure cooperative perception and forecasting},
  author={Yu, Haibao and Yang, Wenxian and Ruan, Hongzhi and Yang, Zhenwei and Tang, Yingjuan and Gao, Xu and Hao, Xin and Shi, Yifeng and Pan, Yifeng and Sun, Ning and Song, Juan and Yuan, Jirui and Luo, Ping and Nie, Zaiqing},
  booktitle={Proceedings of the IEEE/CVF Conference on Computer Vision and Pattern Recognition},
  year={2023},
}

@inproceedings{xiang2024v2x,
  title={V2x-real: a largs-scale dataset for vehicle-to-everything cooperative perception},
  author={Xiang, Hao and Zheng, Zhaoliang and Xia, Xin and Xu, Runsheng and Gao, Letian and Zhou, Zewei and Han, Xu and Ji, Xinkai and Li, Mingxi and Meng, Zonglin and others},
  booktitle={European Conference on Computer Vision},
  pages={455--470},
  year={2024},
  organization={Springer}
}

@inproceedings{yan2024street,
  title={Street gaussians: Modeling dynamic urban scenes with gaussian splatting},
  author={Yan, Yunzhi and Lin, Haotong and Zhou, Chenxu and Wang, Weijie and Sun, Haiyang and Zhan, Kun and Lang, Xianpeng and Zhou, Xiaowei and Peng, Sida},
  booktitle={European Conference on Computer Vision},
  pages={156--173},
  year={2024},
  organization={Springer}
}

@article{chen2026periodic,
  title={Periodic vibration gaussian: Dynamic urban scene reconstruction and real-time rendering},
  author={Chen, Yurui and Gu, Chun and Jiang, Junzhe and Zhu, Xiatian and Zhang, Li},
  journal={International Journal of Computer Vision},
  volume={134},
  number={3},
  pages={83},
  year={2026},
  publisher={Springer}
}

@inproceedings{yang2024deformable,
  title={Deformable 3d gaussians for high-fidelity monocular dynamic scene reconstruction},
  author={Yang, Ziyi and Gao, Xinyu and Zhou, Wen and Jiao, Shaohui and Zhang, Yuqing and Jin, Xiaogang},
  booktitle={Proceedings of the IEEE/CVF conference on computer vision and pattern recognition},
  pages={20331--20341},
  year={2024}
}

@inproceedings{xu2025cruise,
  title={Cruise: Cooperative reconstruction and editing in v2x scenarios using gaussian splatting},
  author={Xu, Haoran and Zhang, Saining and Li, Peishuo and Ye, Baijun and Chen, Xiaoxue and Gao, Huan-ang and Zheng, Jv and Song, Xiaowei and Peng, Ziqiao and Miao, Run and others},
  booktitle={2025 IEEE/RSJ International Conference on Intelligent Robots and Systems (IROS)},
  pages={12518--12525},
  year={2025},
  organization={IEEE}
}

@article{kerbl20233d,
  title={3d gaussian splatting for real-time radiance field rendering.},
  author={Kerbl, Bernhard and Kopanas, Georgios and Leimk{\"u}hler, Thomas and Drettakis, George and others},
  journal={ACM Trans. Graph.},
  volume={42},
  number={4},
  pages={139--1},
  year={2023}
}

@inproceedings{pumarola2021d,
  title={D-nerf: Neural radiance fields for dynamic scenes},
  author={Pumarola, Albert and Corona, Enric and Pons-Moll, Gerard and Moreno-Noguer, Francesc},
  booktitle={Proceedings of the IEEE/CVF conference on computer vision and pattern recognition},
  pages={10318--10327},
  year={2021}
}

@inproceedings{li2021neural,
  title={Neural scene flow fields for space-time view synthesis of dynamic scenes},
  author={Li, Zhengqi and Niklaus, Simon and Snavely, Noah and Wang, Oliver},
  booktitle={Proceedings of the IEEE/CVF conference on computer vision and pattern recognition},
  pages={6498--6508},
  year={2021}
}

@inproceedings{gao2021dynamic,
  title={Dynamic view synthesis from dynamic monocular video},
  author={Gao, Chen and Saraf, Ayush and Kopf, Johannes and Huang, Jia-Bin},
  booktitle={Proceedings of the IEEE/CVF International Conference on Computer Vision},
  pages={5712--5721},
  year={2021}
}

@inproceedings{cao2023hexplane,
  title={Hexplane: A fast representation for dynamic scenes},
  author={Cao, Ang and Johnson, Justin},
  booktitle={Proceedings of the IEEE/CVF Conference on Computer Vision and Pattern Recognition},
  pages={130--141},
  year={2023}
}

@inproceedings{fridovich2023k,
  title={K-planes: Explicit radiance fields in space, time, and appearance},
  author={Fridovich-Keil, Sara and Meanti, Giacomo and Warburg, Frederik Rahb{\ae}k and Recht, Benjamin and Kanazawa, Angjoo},
  booktitle={Proceedings of the IEEE/CVF conference on computer vision and pattern recognition},
  pages={12479--12488},
  year={2023}
}

@inproceedings{wu20244d,
  title={4d gaussian splatting for real-time dynamic scene rendering},
  author={Wu, Guanjun and Yi, Taoran and Fang, Jiemin and Xie, Lingxi and Zhang, Xiaopeng and Wei, Wei and Liu, Wenyu and Tian, Qi and Wang, Xinggang},
  booktitle={Proceedings of the IEEE/CVF conference on computer vision and pattern recognition},
  pages={20310--20320},
  year={2024}
}

@article{li2021learning,
  title={Learning distilled collaboration graph for multi-agent perception},
  author={Li, Yiming and Ren, Shunli and Wu, Pengxiang and Chen, Siheng and Feng, Chen and Zhang, Wenjun},
  journal={Advances in Neural Information Processing Systems},
  volume={34},
  pages={29541--29552},
  year={2021}
}

@inproceedings{xu2022v2x,
  title={V2x-vit: Vehicle-to-everything cooperative perception with vision transformer},
  author={Xu, Runsheng and Xiang, Hao and Tu, Zhengzhong and Xia, Xin and Yang, Ming-Hsuan and Ma, Jiaqi},
  booktitle={European conference on computer vision},
  pages={107--124},
  year={2022},
  organization={Springer}
}

@article{hu2022where2comm,
  title={Where2comm: Communication-efficient collaborative perception via spatial confidence maps},
  author={Hu, Yue and Fang, Shaoheng and Lei, Zixing and Zhong, Yiqi and Chen, Siheng},
  journal={Advances in neural information processing systems},
  volume={35},
  pages={4874--4886},
  year={2022}
}

@article{lu2022robust,
  title={Robust collaborative 3d object detection in presence of pose errors},
  author={Lu, Yifan and Li, Quanhao and Liu, Baoan and Dianati, Mehrdad and Feng, Chen and Chen, Siheng and Wang, Yanfeng},
  journal={arXiv preprint arXiv:2211.07214},
  year={2022}
}

@article{Kuhn,  
 title={The Hungarian method for the assignment problem}, 
 url={http://dx.doi.org/10.1002/nav.3800020109}, 
 DOI={10.1002/nav.3800020109}, 
 journal={Naval Research Logistics Quarterly}, 
 author={Kuhn, H. W.}, 
 pages={83–97}, 
 language={en-US},
 year={1955}
 }

@inproceedings{teed2020raft,
  title={Raft: Recurrent all-pairs field transforms for optical flow},
  author={Teed, Zachary and Deng, Jia},
  booktitle={European conference on computer vision},
  pages={402--419},
  year={2020},
  organization={Springer}
}

@article{unterthiner2018towards,
  title={Towards accurate generative models of video: A new metric \& challenges},
  author={Unterthiner, Thomas and Van Steenkiste, Sjoerd and Kurach, Karol and Marinier, Raphael and Michalski, Marcin and Gelly, Sylvain},
  journal={arXiv preprint arXiv:1812.01717},
  year={2018}
}

@article{yang2023emernerf,
  title={Emernerf: Emergent spatial-temporal scene decomposition via self-supervision},
  author={Yang, Jiawei and Ivanovic, Boris and Litany, Or and Weng, Xinshuo and Kim, Seung Wook and Li, Boyi and Che, Tong and Xu, Danfei and Fidler, Sanja and Pavone, Marco and others},
  journal={arXiv preprint arXiv:2311.02077},
  year={2023}
}

@INPROCEEDINGS{9812038,
  author={Xu, Runsheng and Xiang, Hao and Xia, Xin and Han, Xu and Li, Jinlong and Ma, Jiaqi},
  booktitle={2022 International Conference on Robotics and Automation (ICRA)}, 
  title={OPV2V: An Open Benchmark Dataset and Fusion Pipeline for Perception with Vehicle-to-Vehicle Communication}, 
  year={2022},
  volume={},
  number={},
  pages={2583-2589},
  doi={10.1109/ICRA46639.2022.9812038}}

@ARTICLE{1284395,
  author={Zhou Wang and Bovik, A.C. and Sheikh, H.R. and Simoncelli, E.P.},
  journal={IEEE Transactions on Image Processing}, 
  title={Image quality assessment: from error visibility to structural similarity}, 
  year={2004},
  volume={13},
  number={4},
  pages={600-612},
  doi={10.1109/TIP.2003.819861}}

@inproceedings{zhang2018perceptual,
  title={The Unreasonable Effectiveness of Deep Features as a Perceptual Metric},
  author={Zhang, Richard and Isola, Phillip and Efros, Alexei A and Shechtman, Eli and Wang, Oliver},
  booktitle={CVPR},
  year={2018}
}

\newpage
\appendix

\clearpage
\phantomsection
\addcontentsline{toc}{section}{Appendices}  

\begin{center}
  {\LARGE\bfseries Appendices}\\[0.8em]
  \textcolor{gray}{\rule{0.5\linewidth}{0.4pt}}\\[1.5em]
\end{center}

\startcontents[appendices]
\printcontents[appendices]{}{1}{%
  \setcounter{tocdepth}{2}%
}

\vspace{2em}
\textcolor{gray}{\rule{\linewidth}{0.4pt}}

\label{sec:appendix}

\section{Notation}
\label{app:notation}
{\small

 \begin{longtable}{@{\hspace{6pt}} p{3.6cm} p{9.6cm} @{\hspace{6pt}}}

  \caption{Notation used in the main paper and appendix.}
  \label{tab:full_notation} \\

  \toprule[1.2pt]
  \textit{Notation} & \textit{Definition} \\
  \midrule[0.6pt]
  \endfirsthead

  \multicolumn{2}{c}{\small\tablename~\thetable{} (continued)} \\[2pt]
  \toprule[1.2pt]
  \textit{Notation} & \textit{Definition} \\
  \midrule[0.6pt]
  \endhead

  \bottomrule[1.2pt]
  \endfoot

  \bottomrule[1.2pt]
  \endlastfoot

  \rowcolor{gray!8}
  \multicolumn{2}{@{\hspace{6pt}}l@{\hspace{6pt}}}%
    {\small\textsc{Indices \& Temporal Quantities}} \\[1pt]
  $c \in \{v, f\}$
    & Source index: $v$ for vehicle, $f$ for infrastructure \\
  $n = 1,\dots,N$
    & Index over canonical Gaussians of a single dynamic agent \\
  $t_{i}$
    & $i$-th cooperative anchor timestamp \\
  $\tau_{c}$
    & Actual sensor capture time of source $c$ \\
  $\Delta\tau$
    & Temporal offset between sources: $\Delta\tau = \tau_{v} - \tau_{f}$ \\
  $v \in \mathbb{R}^{3}$
    & Agent velocity vector under locally rigid motion \\[3pt]

  \rowcolor{gray!8}
  \multicolumn{2}{@{\hspace{6pt}}l@{\hspace{6pt}}}%
    {\small\textsc{Canonical-Space Parameters (\textcolor{OrangeVar}{orange})}} \\[1pt]
  $\vo{\bar{\mu}_{n}} \in \mathbb{R}^{3}$
    & Canonical mean of Gaussian $n$ in the agent's body frame \\
  $\vo{\bar{\mu}_{n}^{*}} \in \mathbb{R}^{3}$
    & Ground-truth canonical mean of Gaussian $n$; shared across all sources \\
  $\vo{\bar{\Sigma}_{n}} \in \mathbb{S}_{++}^{3}$,
    \; $\vo{\bar{\alpha}_{n}}$,
    \; $\vo{\bar{c}_{n}}$
    & Canonical covariance, opacity, and color coefficient of Gaussian $n$ \\
  $\vo{w_{n}^{c}(u)}$
    & Alpha-compositing blending weight of Gaussian $n$ at pixel $u$
      from source $c$ \\
  $\vo{\phi_{n}^{c}(u)} \in \mathbb{R}^{3}$
    & Pixel-space influence vector of Gaussian $n$ at pixel $u$ from
      source $c$ (Eq.~\ref{eq:influence_def}) \\
  $\vo{\mathbf{A}_{n}^{c}} \in \mathbb{S}_{+}^{3}$
    & Rendering Fisher information matrix:
      $\vo{\mathbf{A}_{n}^{c}}=\sum_{u\in\mathcal{U}^{c}}
      \vo{\phi_{n}^{c}(u)}\,\vo{\phi_{n}^{c}(u)}^{\!\top}$
      (Def.~\ref{def:fisher}) \\
  $\lambda_{\min,n}$, $\lambda_{n}$
    & Minimum Fisher eigenvalue:
      $\min\!\bigl(\lambda_{\min}(\vo{\mathbf{A}_{n}^{v}}),\,
      \lambda_{\min}(\vo{\mathbf{A}_{n}^{f}})\bigr)$ \\[3pt]

  \rowcolor{gray!8}
  \multicolumn{2}{@{\hspace{6pt}}l@{\hspace{6pt}}}%
    {\small\textsc{Pose \& World-Space Quantities (\textcolor{BlueVar}{blue})}} \\[1pt]
  $\vb{R^{c}} \in SO(3)$,\; $\vb{T^{c}} \in \mathbb{R}^{3}$
    & Rotation and translation of source $c$ at capture time $\tau_{c}$ \\
  $\vb{R_{i}}$,\; $\vb{T_{i}}$
    & Shared rotation and translation in the single-timeline formulation
      at anchor $t_{i}$ \\
  $\mu_{n}^{c} \in \mathbb{R}^{3}$
    & World-space mean of Gaussian $n$ from source $c$:
      $\mu_{n}^{c}=\vb{R^{c}}\vo{\bar{\mu}_{n}}+\vb{T^{c}}$ \\
  $\mu_{n}^{*,c} \in \mathbb{R}^{3}$
    & Ground-truth world position of Gaussian $n$ at capture time $\tau_{c}$ \\
  $\mu_{n}^{\dagger} \in \mathbb{R}^{3}$
    & Fisher-weighted optimal world position under single-timeline
      (Eq.~\ref{eq:opt_mu}) \\
  $\delta_{n}^{c} \in \mathbb{R}^{3}$
    & World-space positional error:
      $\delta_{n}^{c}=\mu_{n}^{c}-\mu_{n}^{*,c}$ \\
  $\delta_{n}^{c\dagger} \in \mathbb{R}^{3}$
    & Residual positional error at the single-timeline optimum
      $\mu_{n}^{\dagger}$ \\
  $\vb{J_{\pi,n}^{c}} \in \mathbb{R}^{2\times 3}$
    & Jacobian of camera projection at $\mu_{n}^{c}$:
      $\vb{J_{\pi,n}^{c}}=\partial\vb{\hat{u}_{n}^{c}}/\partial\mu_{n}^{c}$ \\
  $\vb{\hat{u}_{n}^{c}} \in \mathbb{R}^{2}$,
    \; $\vb{\hat{\Sigma}_{n}^{c}} \in \mathbb{S}_{++}^{2}$
    & Projected 2D mean and covariance of Gaussian $n$ in source $c$'s
      image plane (EWA splatting) \\
  $\vb{\xi^{v}},\vb{\xi^{f}} \in \mathbb{R}^{6}$
    & Source-specific pose parameters in $\mathfrak{se}(3)$ under DUST-GSG;
      collapsed to a single shared $\vb{\xi}$ under single-timeline \\[3pt]

  \rowcolor{gray!8}
  \multicolumn{2}{@{\hspace{6pt}}l@{\hspace{6pt}}}%
    {\small\textsc{NTK Quantities}} \\[1pt]
  $\theta = (\vo{\bar{\boldsymbol{\mu}}},\vb{\boldsymbol{\xi}})$
    & Full optimizable parameter vector \\
  $\vb{\Theta} \in \mathbb{R}^{M\times M}$
    & Empirical Neural Tangent Kernel matrix \\
  $\vb{\Theta}^{\vo{\bar{\boldsymbol{\mu}}}}$,
    \; $\vb{\Theta}^{\vb{\boldsymbol{\xi}}}$
    & Canonical-parameter and pose-parameter additive components of
      $\vb{\Theta}$ \\
  $\vb{\Theta_{vv}^{\xi^{v}}}$,
    \; $\vb{\Theta_{ff}^{\xi^{f}}}$
    & Vehicle and infrastructure diagonal blocks of the DUST-GSG pose kernel \\
  $\mathbf{r}^{c}(t)$
    & Pixel residual vector of source $c$ at training step $t$:
      $\mathbf{r}^{c}(t)=\hat{\mathbf{I}}^{c}(t)-\mathbf{I}^{c*}$ \\[3pt]

  \rowcolor{gray!8}
  \multicolumn{2}{@{\hspace{6pt}}l@{\hspace{6pt}}}%
    {\small\textsc{Rendering \& Photometric Loss}} \\[1pt]
  $\hat{I}^{c}(u)$
    & Rendered color at pixel $u$ from source $c$ \\
  $I^{c*}(u)$
    & Ground-truth image captured at time $\tau_{c}$ \\
  $\mathcal{U}^{c}$
    & Set of pixels observing the agent from source $c$ \\
  $\mathcal{L}^{c}$
    & Per-source photometric loss \\
  $\mathcal{L}_{n}^{c}$
    & Per-Gaussian per-source loss (quadratic form) \\
  $\mathcal{L}_{\text{single}}^{*}$
    & Optimal reconstruction loss under single-timeline \\
  $\mathcal{L}_{\text{DST}}^{*}$
    & Optimal reconstruction loss under DUST \\[3pt]

  \rowcolor{gray!8}
  \multicolumn{2}{@{\hspace{6pt}}l@{\hspace{6pt}}}%
    {\small\textsc{Loss Functions \& Regularization}} \\[1pt]
  $\mathcal{L}_{\text{image}}$
    & Image loss: $(1-\lambda_{r})\mathcal{L}_{1} + \lambda_{r}\mathcal{L}_{\text{SSIM}}$ \\
  $\mathcal{L}_{1}$
    & L1 reconstruction loss \\
  $\mathcal{L}_{\text{SSIM}}$
    & SSIM reconstruction loss \\
  $\lambda_{r}=0.2$
    & Weight of the SSIM loss \\
  $\mathcal{L}_{\text{depth}}$
    & Depth map loss: $\frac{1}{hw}\sum\|\mathcal{D}^{s} - \hat{\mathcal{D}}\|_{1}$ \\
  $\mathcal{D}^{s}$
    & Inverse of the sparse LiDAR depth map \\
  $\hat{\mathcal{D}}$
    & Inverse of the predicted depth map \\
  $\mathcal{L}_{\text{opacity}}$
    & Mask/opacity loss (sky + occupancy) \\
  $O_{g}$
    & Rendered opacity map \\
  $M_{\text{sky}}$
    & Sky mask \\
  $\mathcal{L}_{\text{reg}}$
    & Pose smoothness regularization \\
  $\boldsymbol{\theta}(t)$
    & Human body poses at time $t$ \\
  $\delta$
    & Randomly chosen integer from $\{1,2,3,4,5\}$ for pose regularization \\
  $\lambda_{\text{depth}}=0.01$
    & Weight of depth loss \\
  $\lambda_{\text{opacity}}=0.05$
    & Weight of opacity loss \\
  $\lambda_{\text{reg}}=0.01$
    & Weight of pose smoothness loss \\[3pt]

  \rowcolor{gray!8}
  \multicolumn{2}{@{\hspace{6pt}}l@{\hspace{6pt}}}%
    {\small\textsc{Evaluation Metrics}} \\[1pt]
  $\hat{\mathbf{I}}$
    & Rendered image \\
  $\mathbf{I}$
    & Ground-truth image \\
  $N$
    & Number of valid pixels (in PSNR computation) \\
  $\mu_{x},\, \mu_{y}$
    & Local means of image patches $\mathbf{x},\mathbf{y}$ (SSIM) \\
  $\sigma_{x},\, \sigma_{y}$
    & Local standard deviations (SSIM) \\
  $\sigma_{xy}$
    & Local cross-covariance (SSIM) \\
  $c_{1},\, c_{2}$
    & Small constants for numerical stability (SSIM) \\
  $\phi_{\ell}(\cdot)$
    & Feature map from layer $\ell$ of AlexNet (LPIPS) \\
  $\mathbf{w}_{\ell}$
    & Learned channel-wise weights at layer $\ell$ (LPIPS) \\
  $H_{\ell},\, W_{\ell}$
    & Spatial dimensions of feature map at layer $\ell$ (LPIPS) \\
  $\boldsymbol{\mu}_{\text{gen}},\, \boldsymbol{\Sigma}_{\text{gen}}$
    & Mean and covariance of generated video clip features (FVD) \\
  $\boldsymbol{\mu}_{\text{real}},\, \boldsymbol{\Sigma}_{\text{real}}$
    & Mean and covariance of real video clip features (FVD) \\
  $T$
    & Total number of video frames \\
  $T_{c}=16$
    & Length of each non-overlapping video clip (FVD) \\
  $\mathbf{F}_{t} \in \mathbb{R}^{H\times W\times 2}$
    & Dense optical flow field between frames $t$ and $t+1$ \\
  $(u,v)$
    & Horizontal and vertical components of optical flow \\
  $(\hat{u}, \hat{v})$
    & Predicted flow components (rendered video) \\
  $\mathcal{M}$
    & Union of dynamic masks from two consecutive frames (EPE) \\

  \end{longtable}

\section{Proof}
\label{app:proof_gradient}

We provide a rigorous proof that the DUST-GSG formulation eliminates the
fundamental optimization conflict inherent in single-timeline
representations under asynchronous vehicle--infrastructure observations.
The proof proceeds in five stages: we first formalize the rendering model
and state the working assumptions (\S\ref{app:model}); then derive the
Jacobian structure and the rendering Fisher information matrix
(\S\ref{app:jacobian}); next prove the irreducible-residual lower bound
(\S\ref{app:part_i}); then establish gradient incompatibility
(\S\ref{app:part_ii}); and finally show how DUST-GSG resolves both issues
via NTK block-diagonalization (\S\ref{app:part_iii}).

\subsection{Formal Rendering Model and Assumptions}
\label{app:model}

Throughout the proof, \vo{orange} denotes \emph{canonical-space} quantities
(those defined in the agent's body frame, independent of capture time), and
\vb{blue} denotes \emph{pose/world-space} quantities (those that depend on
the source-specific rigid-body transform or the camera).

\begin{definition}[Canonical Gaussian Agent]
\label{def:canonical}
A dynamic agent is represented by $N$ canonical 3D Gaussians with means
$\vo{\bar{\bm{\mu}}}=(\bar{\mu}_{1},\dots,\bar{\mu}_{N})\in\mathbb{R}^{3N}$,
covariances $\vo{\bar{\bm{\Sigma}}}=(\bar{\Sigma}_{1},\dots,\bar{\Sigma}_{N})$
with each $\bar{\Sigma}_{n}\in\mathbb{S}_{++}^{3}$, opacities
$\vo{\bar{\bm{\alpha}}}=(\bar{\alpha}_{1},\dots,\bar{\alpha}_{N})\in(0,1)^{N}$,
and view-dependent color coefficients
$\vo{\bar{\mathbf{c}}}=(\bar{c}_{1},\dots,\bar{c}_{N})$.
\end{definition}

\begin{definition}[Source-Specific World Projection]
\label{def:world_proj}
For source $c\in\{v,f\}$ (vehicle, infrastructure) with rigid-body pose
$\vb{P^{c}}=(\vb{R^{c}},\vb{T^{c}})\in SE(3)$ at capture time $\tau_{c}$,
the world-space parameters of Gaussian $n$ are
\begin{equation}
\label{eq:world_transform}
  \mu_{n}^{c} = \vb{R^{c}}\,\vo{\bar{\mu}_{n}} + \vb{T^{c}},
  \qquad
  \Sigma_{n}^{c} = \vb{R^{c}}\,\vo{\bar{\Sigma}_{n}}\,(\vb{R^{c}})^{\!\top}.
\end{equation}
\end{definition}

\begin{definition}[Differentiable Splatting Renderer]
\label{def:renderer}
Given camera intrinsics $\vb{K^{c}}\!\in\!\mathbb{R}^{3\times 3}$ and
extrinsics $\vb{W^{c}}\!\in\!SE(3)$, the rendered color at pixel
$u\in\mathbb{R}^{2}$ from source $c$ is
\begin{equation}
\label{eq:rendering}
  \hat{I}^{c}(u)
  =\sum_{n=1}^{N}
  \underbrace{
    \vo{\bar{\alpha}_{n}}
    \prod_{m<n}\!\bigl(1-\vo{\bar{\alpha}_{m}}\,\vb{\mathcal{G}_{m}^{c}(u)}\bigr)
  }_{\displaystyle\vo{w_{n}^{c}(u)}}
  \;\vo{\bar{c}_{n}}\;\vb{\mathcal{G}_{n}^{c}(u)},
\end{equation}
where $\vb{\mathcal{G}_{n}^{c}(u)}
=\exp\!\bigl(-\tfrac{1}{2}(u-\vb{\hat{u}_{n}^{c}})^{\!\top}
(\vb{\hat{\Sigma}_{n}^{c}})^{-1}(u-\vb{\hat{u}_{n}^{c}})\bigr)$
is the 2D Gaussian evaluated at pixel $u$, with projected mean
$\vb{\hat{u}_{n}^{c}}=\pi(\vb{K^{c}W^{c}}\mu_{n}^{c})$ and projected
covariance $\vb{\hat{\Sigma}_{n}^{c}}$ obtained via EWA splatting.
The blending weight $\vo{w_{n}^{c}(u)}$ encapsulates the front-to-back
alpha-compositing logic.
\end{definition}

\begin{assumption}[Linearized Rendering Regime]
\label{assump:linear}
We approximate the rendering function by its first-order Taylor expansion
around the current parameters $\theta_{0}$. This characterizes the early training phase where gradient conflicts are
most damaging:
\begin{equation}
\label{eq:linearized}
  \hat{I}^{c}(u;\theta)
  \;\approx\;
  \hat{I}^{c}(u;\theta_{0})
  +
  \bigl\langle
    \nabla_{\!\theta}\hat{I}^{c}(u;\theta_{0}),\;
    \theta-\theta_{0}
  \bigr\rangle.
\end{equation}
\end{assumption}

\begin{assumption}[Locally Rigid Motion]
\label{assump:rigid}
During the temporal offset $\Delta\tau=\tau_{v}-\tau_{f}$ between the two
sources, the agent undergoes approximately constant-velocity motion with
velocity $v\in\mathbb{R}^{3}$:
\begin{equation}
\label{eq:rigid_motion}
  \mu_{n}^{*}(\tau_{f})
  \;\approx\;
  \mu_{n}^{*}(\tau_{v}) + v\cdot\Delta\tau,
  \qquad\forall\;n=1,\dots,N,
\end{equation}
where $\mu_{n}^{*}(\tau_{c})$ denotes the ground-truth world position of
Gaussian $n$ at time $\tau_{c}$.
\end{assumption}

\subsection{Jacobian Structure and Rendering Fisher Information}
\label{app:jacobian}

From the rendering equation~\eqref{eq:rendering}, the gradient of
$\hat{I}^{c}(u)$ with respect to $\vo{\bar{\mu}_{n}}$ factors through
three stages via the chain rule:
\begin{equation}
\label{eq:chain_rule}
  \nabla_{\!\vo{\bar{\mu}_{n}}}\hat{I}^{c}(u)
  \;=\;
  \underbrace{(\vb{R^{c}})^{\!\top}}_{\text{world}\rightarrow\text{canonical}}
  \;\cdot\;
  \underbrace{(\vb{J_{\pi,n}^{c}})^{\!\top}}_{\text{image}\rightarrow\text{world}}
  \;\cdot\;
  \underbrace{
    \nabla_{\!\vb{\hat{u}_{n}^{c}}}\hat{I}^{c}(u)
  }_{\text{color}\rightarrow\text{image}}
  \;\;\in\mathbb{R}^{3},
\end{equation}

where $\vb{J_{\pi,n}^{c}}=\frac{\partial\vb{\hat{u}_{n}^{c}}}{\partial\mu_{n}^{c}}\in\mathbb{R}^{2\times 3}$
is the Jacobian of the camera projection evaluated at $\mu_{n}^{c}$, and
$\nabla_{\!\vb{\hat{u}_{n}^{c}}}\hat{I}^{c}(u)\in\mathbb{R}^{2}$ is the
sensitivity of the rendered color to the projected Gaussian center.

Differentiating Eq.~\eqref{eq:rendering} with respect to
$\vb{\hat{u}_{n}^{c}}$:
\begin{equation}
\label{eq:pixel_sensitivity}
  \nabla_{\!\vb{\hat{u}_{n}^{c}}}\hat{I}^{c}(u)
  \;\approx\;
  \vo{w_{n}^{c}(u)}\;\vo{\bar{c}_{n}}\;\vb{\mathcal{G}_{n}^{c}(u)}\;
  (\vb{\hat{\Sigma}_{n}^{c}})^{-1}(u-\vb{\hat{u}_{n}^{c}})
  \;\;\in\mathbb{R}^{2}.
\end{equation}
The sensitivity is proportional to the blending weight $\vo{w_{n}^{c}(u)}$,
the Gaussian kernel value $\vb{\mathcal{G}_{n}^{c}(u)}$, and the Mahalanobis
displacement $(u-\vb{\hat{u}_{n}^{c}})$ scaled by the inverse projected
covariance. Pixels far from the Gaussian center or occluded by other
Gaussians contribute negligibly.

Substituting Eq.~\eqref{eq:pixel_sensitivity} into Eq.~\eqref{eq:chain_rule}
and defining the \emph{pixel-space influence vector}
\begin{equation}
\label{eq:influence_def}
  \vo{\phi_{n}^{c}(u)}
  \;\coloneqq\;
  (\vb{J_{\pi,n}^{c}})^{\!\top}\;
  \vo{w_{n}^{c}(u)}\;\vo{\bar{c}_{n}}\;\vb{\mathcal{G}_{n}^{c}(u)}\;
  (\vb{\hat{\Sigma}_{n}^{c}})^{-1}(u-\vb{\hat{u}_{n}^{c}})
  \;\;\in\mathbb{R}^{3},
\end{equation}
we obtain the compact form
\begin{equation}
\label{eq:grad_compact}
  \nabla_{\!\vo{\bar{\mu}_{n}}}\hat{I}^{c}(u)
  \;=\;
  (\vb{R^{c}})^{\!\top}\;\vo{\phi_{n}^{c}(u)}.
\end{equation}
By Eq.~\eqref{eq:grad_compact}, the gradient in canonical space equals the
world-space influence vector $\vo{\phi_{n}^{c}(u)}$ rotated back by
$(\vb{R^{c}})^{\!\top}$. This factorization is the structural key of the
entire proof: since $\vb{R^{v}}\neq\vb{R^{f}}$ in general, the same
canonical parameter $\vo{\bar{\mu}_{n}}$ is pulled in two
\emph{different rotated directions} by the two sources---the geometric
root of the gradient conflict exposed in Part~(ii).

Consider the per-source photometric loss
$\mathcal{L}^{c}=\frac{1}{2}\sum_{u\in\mathcal{U}^{c}}
(\hat{I}^{c}(u)-I^{c*}(u))^{2}$,
where $\mathcal{U}^{c}$ is the set of pixels observing the agent from
source $c$ and $I^{c*}$ is the ground-truth image captured at time
$\tau_{c}$. The gradient with respect to $\vo{\bar{\mu}_{n}}$ is
\begin{equation}
\label{eq:grad_loss}
  \nabla_{\!\vo{\bar{\mu}_{n}}}\mathcal{L}^{c}
  =
  (\vb{R^{c}})^{\!\top}
  \sum_{u\in\mathcal{U}^{c}}
  \bigl(\hat{I}^{c}(u)-I^{c*}(u)\bigr)\;
  \vo{\phi_{n}^{c}(u)}.
\end{equation}
Under Assumption~\ref{assump:linear}, the photometric residual at pixel $u$
is dominated by the world-space positional error
$\delta_{n}^{c}=\mu_{n}^{c}-\mu_{n}^{*,c}$ of Gaussian $n$. A first-order
expansion of the rendering equation around the true position yields
$\hat{I}^{c}(u)-I^{c*}(u)\approx\vo{\phi_{n}^{c}(u)}^{\!\top}\delta_{n}^{c}$
(contribution from Gaussian $n$). Substituting into
Eq.~\eqref{eq:grad_loss}, the gradient becomes
\begin{equation}
\label{eq:grad_quadratic}
  \nabla_{\!\vo{\bar{\mu}_{n}}}\mathcal{L}^{c}
  \;\approx\;
  (\vb{R^{c}})^{\!\top}\;
  \underbrace{
    \left(\sum_{u\in\mathcal{U}^{c}}
      \vo{\phi_{n}^{c}(u)}\;\vo{\phi_{n}^{c}(u)}^{\!\top}
    \right)
  }_{\displaystyle\vo{\mathbf{A}_{n}^{c}}\;\in\;\mathbb{R}^{3\times 3}}
  \;\delta_{n}^{c}.
\end{equation}

\begin{definition}[Rendering Fisher Information Matrix]
\label{def:fisher}
The \emph{rendering Fisher information matrix} of Gaussian $n$ from source
$c$ is
\begin{equation}
\label{eq:fisher_def}
  \vo{\mathbf{A}_{n}^{c}}
  \;=\;
  \sum_{u\in\mathcal{U}^{c}}
  \vo{\phi_{n}^{c}(u)}\;\vo{\phi_{n}^{c}(u)}^{\!\top}
  \;\;\in\;\mathbb{S}_{+}^{3}.
\end{equation}
\end{definition}

Being a sum of rank-1 outer products, $\vo{\mathbf{A}_{n}^{c}}$ is positive
semi-definite by construction. It becomes strictly positive definite
($\vo{\mathbf{A}_{n}^{c}}\succ 0$) whenever
$\{\vo{\phi_{n}^{c}(u)}\}_{u\in\mathcal{U}^{c}}$ spans $\mathbb{R}^{3}$,
which holds when the agent is observed from a non-degenerate viewpoint with
sufficient pixel coverage. We assume $\vo{\mathbf{A}_{n}^{c}}\succ 0$
throughout. Combining Eq.~\eqref{eq:fisher_def} with the residual
approximation, the per-source loss for Gaussian $n$ takes the quadratic form
\begin{equation}
\label{eq:loss_quadratic}
  \mathcal{L}_{n}^{c}
  \;\approx\;
  \frac{1}{2}\;
  (\delta_{n}^{c})^{\!\top}\;\vo{\mathbf{A}_{n}^{c}}\;\delta_{n}^{c},
\end{equation}
and the total per-source loss is $\mathcal{L}^{c}=\sum_{n=1}^{N}\mathcal{L}_{n}^{c}$.
This quadratic structure is the foundation for all subsequent analyses.

\subsection{Proof of Part (i): Irreducible Residual under Single-Timeline}
\label{app:part_i}

In the single-timeline formulation, both sources share pose
$\vb{P(t_{i})}=(\vb{R_{i}},\vb{T_{i}})$ at the anchor timestamp $t_{i}$,
so the world-space position of every Gaussian is the same for both sources:
$\mu_{n}^{v}=\mu_{n}^{f}=\vb{R_{i}}\,\vo{\bar{\mu}_{n}}+\vb{T_{i}}$.
However, the ground-truth positions differ between the two capture times.
Under Assumption~\ref{assump:rigid}:
\begin{equation}
\label{eq:true_pos}
  \mu_{n}^{*,v}=\mu_{n}^{*}(t_{i})+v(\tau_{v}-t_{i}),
  \qquad
  \mu_{n}^{*,f}=\mu_{n}^{*}(t_{i})+v(\tau_{f}-t_{i}).
\end{equation}
Denoting the shared world position as
$\mu_{n}=\vb{R_{i}}\vo{\bar{\mu}_{n}}+\vb{T_{i}}$,
the positional errors for the two sources are
\begin{align}
\label{eq:pos_errors}
  \delta_{n}^{v}
  &=\mu_{n}-\mu_{n}^{*,v}
   =\bigl(\mu_{n}-\mu_{n}^{*}(t_{i})\bigr)-v(\tau_{v}-t_{i}),
  \nonumber\\
  \delta_{n}^{f}
  &=\mu_{n}-\mu_{n}^{*,f}
   =\bigl(\mu_{n}-\mu_{n}^{*}(t_{i})\bigr)-v(\tau_{f}-t_{i}).
\end{align}
Crucially, the difference between the two errors is
\emph{independent of the parameters}:
\begin{equation}
\label{eq:error_diff}
  \delta_{n}^{v}-\delta_{n}^{f}
  \;=\;
  -v\,\Delta\tau,
  \qquad\Delta\tau=\tau_{v}-\tau_{f}.
\end{equation}
This identity is the root cause of the optimization conflict: no matter how
we adjust $\vo{\bar{\mu}_{n}}$, the two positional errors always differ by
the fixed vector $-v\Delta\tau$. The optimizer therefore faces an
irreconcilable constraint---reducing $\delta_{n}^{v}$ necessarily increases
$\delta_{n}^{f}$ by the same amount, and vice versa.

The total loss for Gaussian $n$ under single-timeline is
$\mathcal{L}_{n}=\mathcal{L}_{n}^{v}+\mathcal{L}_{n}^{f}
=\frac{1}{2}(\delta_{n}^{v})^{\!\top}\vo{\mathbf{A}_{n}^{v}}\delta_{n}^{v}
+\frac{1}{2}(\delta_{n}^{f})^{\!\top}\vo{\mathbf{A}_{n}^{f}}\delta_{n}^{f}$.
We minimize over $\mu_{n}\in\mathbb{R}^{3}$ by setting
$\nabla_{\!\mu_{n}}\mathcal{L}_{n}
=\vo{\mathbf{A}_{n}^{v}}\delta_{n}^{v}+\vo{\mathbf{A}_{n}^{f}}\delta_{n}^{f}
=\mathbf{0}$,
and solving for the optimal position:
\begin{align}
\label{eq:opt_mu}
  \vo{\mathbf{A}_{n}^{v}}\,(\mu_{n}-\mu_{n}^{*,v})
  +\vo{\mathbf{A}_{n}^{f}}\,(\mu_{n}-\mu_{n}^{*,f})
  &=\mathbf{0}
  \nonumber\\
  \bigl(\vo{\mathbf{A}_{n}^{v}}+\vo{\mathbf{A}_{n}^{f}}\bigr)\mu_{n}
  &=\vo{\mathbf{A}_{n}^{v}}\,\mu_{n}^{*,v}
   +\vo{\mathbf{A}_{n}^{f}}\,\mu_{n}^{*,f}
  \nonumber\\
  \mu_{n}^{\dagger}
  &=\bigl(\vo{\mathbf{A}_{n}^{v}}+\vo{\mathbf{A}_{n}^{f}}\bigr)^{-1}
   \bigl(\vo{\mathbf{A}_{n}^{v}}\,\mu_{n}^{*,v}
         +\vo{\mathbf{A}_{n}^{f}}\,\mu_{n}^{*,f}\bigr).
\end{align}
This is the Fisher-weighted average of the two target positions---a
compromise that is optimal for neither source individually. Substituting
$\mu_{n}^{\dagger}$ back, the residual errors at the optimum are
\begin{align}
\label{eq:opt_errors}
  \delta_{n}^{v\dagger}
  &=\mu_{n}^{\dagger}-\mu_{n}^{*,v}
  \nonumber\\
  &=\bigl(\vo{\mathbf{A}_{n}^{v}}+\vo{\mathbf{A}_{n}^{f}}\bigr)^{-1}
    \bigl(\vo{\mathbf{A}_{n}^{v}}\,\mu_{n}^{*,v}
          +\vo{\mathbf{A}_{n}^{f}}\,\mu_{n}^{*,f}\bigr)
    -\mu_{n}^{*,v}
  \nonumber\\
  &=\bigl(\vo{\mathbf{A}_{n}^{v}}+\vo{\mathbf{A}_{n}^{f}}\bigr)^{-1}
    \vo{\mathbf{A}_{n}^{f}}\,(\mu_{n}^{*,f}-\mu_{n}^{*,v})
  \nonumber\\
  &=\bigl(\vo{\mathbf{A}_{n}^{v}}+\vo{\mathbf{A}_{n}^{f}}\bigr)^{-1}
    \vo{\mathbf{A}_{n}^{f}}\,(-v\Delta\tau),
\end{align}
where the last step uses $\mu_{n}^{*,f}-\mu_{n}^{*,v}=-v\Delta\tau$ from
Eqs.~\eqref{eq:true_pos}--\eqref{eq:error_diff}. Similarly,
$\delta_{n}^{f\dagger}
=(\vo{\mathbf{A}_{n}^{v}}+\vo{\mathbf{A}_{n}^{f}})^{-1}
 \vo{\mathbf{A}_{n}^{v}}(v\Delta\tau)$.
Both residual errors are non-zero whenever $v\Delta\tau\neq\mathbf{0}$,
confirming that the Fisher-weighted compromise leaves both sources with
unresolved positional error.

We now derive the lower bound in two steps.

\textit{Step A (eigenvalue inequality).}
For any $M\succ 0$ and any vector $x$,
$x^{\!\top}Mx\geq\lambda_{\min}(M)\|x\|^{2}$. Applying this at the optimum:
\begin{align}
\label{eq:lower_step_a}
  \mathcal{L}_{n}^{*}
  &=\frac{1}{2}(\delta_{n}^{v\dagger})^{\!\top}
    \vo{\mathbf{A}_{n}^{v}}\,\delta_{n}^{v\dagger}
   +\frac{1}{2}(\delta_{n}^{f\dagger})^{\!\top}
    \vo{\mathbf{A}_{n}^{f}}\,\delta_{n}^{f\dagger}
  \nonumber\\
  &\geq
   \frac{\lambda_{\min}(\vo{\mathbf{A}_{n}^{v}})}{2}
   \|\delta_{n}^{v\dagger}\|^{2}
   +
   \frac{\lambda_{\min}(\vo{\mathbf{A}_{n}^{f}})}{2}
   \|\delta_{n}^{f\dagger}\|^{2}
  \nonumber\\
  &\geq
   \frac{\lambda_{\min,n}}{2}
   \bigl(\|\delta_{n}^{v\dagger}\|^{2}+\|\delta_{n}^{f\dagger}\|^{2}\bigr),
\end{align}
where
$\lambda_{\min,n}=\min\!\bigl(\lambda_{\min}(\vo{\mathbf{A}_{n}^{v}}),
\lambda_{\min}(\vo{\mathbf{A}_{n}^{f}})\bigr)>0$.

\textit{Step B (parallelogram identity).}
For any two vectors $a,b\in\mathbb{R}^{3}$,
$\|a\|^{2}+\|b\|^{2}=2\|{(a+b)}/{2}\|^{2}+{\|a-b\|^{2}}/{2}
\geq{\|a-b\|^{2}}/{2}$.
Setting $a=\delta_{n}^{v\dagger}$, $b=\delta_{n}^{f\dagger}$, and using
$\delta_{n}^{v\dagger}-\delta_{n}^{f\dagger}=-v\Delta\tau$:
\begin{equation}
\label{eq:lower_step_b}
  \|\delta_{n}^{v\dagger}\|^{2}+\|\delta_{n}^{f\dagger}\|^{2}
  \;\geq\;
  \frac{\|\delta_{n}^{v\dagger}-\delta_{n}^{f\dagger}\|^{2}}{2}
  =
  \frac{\|v\Delta\tau\|^{2}}{2}
  =
  \frac{\|v\|^{2}|\Delta\tau|^{2}}{2}.
\end{equation}

\textit{Combining Steps A and B} and summing over all $N$ Gaussians:
\begin{equation}
\label{eq:total_lower_bound}
    \mathcal{L}_{\mathrm{single}}^{*}
    =\sum_{n=1}^{N}\mathcal{L}_{n}^{*}
    \;\geq\;
    \frac{\|v\|^{2}|\Delta\tau|^{2}}{4}
    \sum_{n=1}^{N}\lambda_{\min,n}
    \;>\;0.
\end{equation}
This completes the proof of Part~(i). The bound is strict and
\emph{representation-level}: no choice of canonical parameters can reduce
the photometric loss below this threshold, regardless of the optimizer
used. The irreducible error scales quadratically with both the agent
velocity $\|v\|$ and the temporal offset $|\Delta\tau|$, and linearly with
the aggregate Fisher information $\sum_{n}\lambda_{\min,n}$---all
quantities that are large in typical high-speed VICAD scenarios.

The formula for the theorem~\ref{thm:gradient_decoupling} in the Sec~\ref{sec:theory} is obtained by reformulating $\lambda_{\min,n}$ as $\lambda_{n}$.

\begin{equation}
    \mathcal{L}_{\mathrm{single}}^{*}
    \;\ge\;
    \frac{|\Delta\tau|^{2}\|v\|^{2}}{4}
    \sum_{n=1}^{N}\lambda_{n}
    \;>\;0,
\end{equation}

where $\lambda_{n}>0$ is the minimum Fisher-information eigenvalue
of $\mathcal{G}_n$, quantifying its photometric sensitivity to displacement.

\subsection{Proof of Part (ii): Gradient Incompatibility}
\label{app:part_ii}

We now show that the two per-source gradients cannot simultaneously vanish
at any canonical parameter, and quantify the separation between their
individual optima. From Eq.~\eqref{eq:grad_quadratic}, the gradient of
$\mathcal{L}^{c}$ with respect to $\vo{\bar{\mu}_{n}}$ vanishes when
\begin{align}
\label{eq:per_source_opt}
  \nabla_{\!\vo{\bar{\mu}_{n}}}\mathcal{L}^{c}=\mathbf{0}
  &\;\Longleftrightarrow\;
   (\vb{R_{i}})^{\!\top}\vo{\mathbf{A}_{n}^{c}}\,\delta_{n}^{c}=\mathbf{0}
  \nonumber\\
  &\;\Longleftrightarrow\;
   \vo{\mathbf{A}_{n}^{c}}\,\delta_{n}^{c}=\mathbf{0}
  \nonumber\\
  &\;\Longleftrightarrow\;
   \delta_{n}^{c}=\mathbf{0}
  \nonumber\\
  &\;\Longleftrightarrow\;
   \mu_{n}=\mu_{n}^{*,c},
\end{align}
where the second equivalence uses $\vb{R_{i}}\in SO(3)$ (invertible) and
the third uses $\vo{\mathbf{A}_{n}^{c}}\succ 0$ (invertible). Thus
$\mathcal{L}^{v}$ is minimized at $\mu_{n}=\mu_{n}^{*,v}$ and
$\mathcal{L}^{f}$ is minimized at $\mu_{n}=\mu_{n}^{*,f}$.

Since $\mu_{n}=\vb{R_{i}}\vo{\bar{\mu}_{n}}+\vb{T_{i}}$ is a bijection in
$\vo{\bar{\mu}_{n}}$, the per-source canonical optima are
$\vo{\bar{\mu}_{n}^{*,v}}=\vb{R_{i}}^{\!\top}(\mu_{n}^{*,v}-\vb{T_{i}})$
and
$\vo{\bar{\mu}_{n}^{*,f}}=\vb{R_{i}}^{\!\top}(\mu_{n}^{*,f}-\vb{T_{i}})$.
Their separation is
\begin{align}
\label{eq:optima_separation}
  \bigl\|\vo{\bar{\mu}_{n}^{*,v}}-\vo{\bar{\mu}_{n}^{*,f}}\bigr\|
  &=\bigl\|\vb{R_{i}}^{\!\top}(\mu_{n}^{*,v}-\mu_{n}^{*,f})\bigr\|
  \nonumber\\
  &=\bigl\|\mu_{n}^{*,v}-\mu_{n}^{*,f}\bigr\|
  \qquad(\vb{R_{i}}\in SO(3)\text{ is isometric})
  \nonumber\\
  &=\|v\,\Delta\tau\|
  \;=\;\|v\|\,|\Delta\tau|.
\end{align}
Since $v\neq\mathbf{0}$ and $\Delta\tau\neq 0$, this separation is
strictly positive, so no single $\vo{\bar{\mu}_{n}}$ can simultaneously
satisfy both $\nabla_{\!\vo{\bar{\mu}_{n}}}\mathcal{L}^{v}=\mathbf{0}$ and
$\nabla_{\!\vo{\bar{\mu}_{n}}}\mathcal{L}^{f}=\mathbf{0}$. The separation
$\|v\|\,|\Delta\tau|$ is the physical displacement of the agent between the
two capture times projected into canonical space. For a vehicle traveling
at $\|v\|=10\;\mathrm{m/s}$ with $|\Delta\tau|=70\;\mathrm{ms}$, this
equals $0.7\;\mathrm{m}$---far exceeding the typical Gaussian kernel radius
in driving scenes. The optimizer is forced to place $\vo{\bar{\mu}_{n}}$ at
the Fisher-weighted compromise (Eq.~\eqref{eq:opt_mu}), creating blurred,
ghosted reconstructions that satisfy neither view.

\paragraph{Gradient anti-alignment in the worst case.}
When $t_{i}$ lies between the two capture times, i.e.,
$(\tau_{v}-t_{i})(\tau_{f}-t_{i})<0$, the positional offsets
$\delta_{n}^{v}$ and $\delta_{n}^{f}$ at the anchor position
$\mu_{n}=\mu_{n}^{*}(t_{i})$ point in opposite directions along $v$.
Under the isotropic Fisher approximation
$\vo{\mathbf{A}_{n}^{v}}\approx\vo{\mathbf{A}_{n}^{f}}\approx\lambda_{n}I_{3}$,
the per-source gradients from Eq.~\eqref{eq:grad_quadratic} become
\begin{align}
\label{eq:anti_aligned}
  \nabla_{\!\vo{\bar{\mu}_{n}}}\mathcal{L}^{v}
  &\approx
   (\vb{R_{i}})^{\!\top}\vo{\mathbf{A}_{n}^{v}}\,\delta_{n}^{v}
  \nonumber\\
  &=\lambda_{n}(\vb{R_{i}})^{\!\top}
    \bigl(\mu_{n}^{*}(t_{i})-\mu_{n}^{*,v}\bigr)
  \nonumber\\
  &=\lambda_{n}(\vb{R_{i}})^{\!\top}
    \bigl(-v(\tau_{v}-t_{i})\bigr)
   \quad\text{[from Eq.~\eqref{eq:true_pos}]}
  \nonumber\\
  &=-\lambda_{n}(\vb{R_{i}})^{\!\top}v(\tau_{v}-t_{i}),
\end{align}
and analogously
$\nabla_{\!\vo{\bar{\mu}_{n}}}\mathcal{L}^{f}
\approx-\lambda_{n}(\vb{R_{i}})^{\!\top}v(\tau_{f}-t_{i})$.
The cosine similarity between the two gradient vectors is
\begin{align}
\label{eq:cos_minus_one}
  \cos\angle\!\bigl(
    \nabla_{\!\vo{\bar{\bm{\mu}}}}\mathcal{L}^{v},\;
    \nabla_{\!\vo{\bar{\bm{\mu}}}}\mathcal{L}^{f}
  \bigr)
  &=
  \frac{
    \nabla_{\!\vo{\bar{\mu}_{n}}}\mathcal{L}^{v}
    \cdot
    \nabla_{\!\vo{\bar{\mu}_{n}}}\mathcal{L}^{f}
  }{
    \bigl\|\nabla_{\!\vo{\bar{\mu}_{n}}}\mathcal{L}^{v}\bigr\|
    \bigl\|\nabla_{\!\vo{\bar{\mu}_{n}}}\mathcal{L}^{f}\bigr\|
  }
  \nonumber\\
  &=
  \frac{
    \lambda_{n}^{2}
    \bigl\|(\vb{R_{i}})^{\!\top}v\bigr\|^{2}
    (\tau_{v}-t_{i})(\tau_{f}-t_{i})
  }{
    \lambda_{n}^{2}
    \bigl\|(\vb{R_{i}})^{\!\top}v\bigr\|^{2}
    |\tau_{v}-t_{i}|\,|\tau_{f}-t_{i}|
  }
  \nonumber\\
  &=
  \frac{(\tau_{v}-t_{i})(\tau_{f}-t_{i})}{|\tau_{v}-t_{i}|\,|\tau_{f}-t_{i}|}
  \;=\; -1,
\end{align}
where the last equality holds because $(\tau_{v}-t_{i})$ and
$(\tau_{f}-t_{i})$ have opposite signs, and
$\|(\vb{R_{i}})^{\!\top}v\|=\|v\|>0$ since $\vb{R_{i}}\in SO(3)$ is
isometric. This is the most severe form of gradient conflict: the two
sources provide diametrically opposed update signals to the shared
canonical parameters. Even when the anchor does not lie between the capture
times, the cosine similarity satisfies
\begin{equation}
\label{eq:cos_general}
  \cos\angle\!\bigl(
    \nabla_{\!\vo{\bar{\bm{\mu}}}}\mathcal{L}^{v},\;
    \nabla_{\!\vo{\bar{\bm{\mu}}}}\mathcal{L}^{f}
  \bigr)
  =
  \frac{(\tau_{v}-t_{i})(\tau_{f}-t_{i})}{|\tau_{v}-t_{i}|\,|\tau_{f}-t_{i}|}
  < +1
  \quad\text{whenever}\;\Delta\tau\neq 0,
\end{equation}
ensuring persistent gradient interference regardless of anchor placement.
\hfill$\square$

\subsection{Proof of Part (iii): NTK Block-Diagonalization}
\label{app:part_iii}

We now show that DUST-GSG simultaneously eliminates the irreducible residual
and the gradient conflict by introducing source-specific pose degrees of
freedom.

Under DUST-GSG, source $c$ uses its own pose
$\vb{P^{c}(\tau_{c})}=(\vb{R^{c}},\vb{T^{c}})$ at its actual capture time
$\tau_{c}$, so the world-space position of Gaussian $n$ is
$\mu_{n}^{c}=\vb{R^{c}}\,\vo{\bar{\mu}_{n}}+\vb{T^{c}}$.
When the poses match the ground truth, the positional error for source $c$
becomes
\begin{align}
\label{eq:dst_error}
  \delta_{n}^{c}
  &=\mu_{n}^{c}-\mu_{n}^{*,c}
  \nonumber\\
  &=\vb{R^{*}(\tau_{c})}\,\vo{\bar{\mu}_{n}}+\vb{T^{*}(\tau_{c})}
   -\bigl(\vb{R^{*}(\tau_{c})}\,\vo{\bar{\mu}_{n}^{*}}+\vb{T^{*}(\tau_{c})}\bigr)
  \nonumber\\
  &=\vb{R^{*}(\tau_{c})}\bigl(\vo{\bar{\mu}_{n}}-\vo{\bar{\mu}_{n}^{*}}\bigr),
\end{align}
where $\vo{\bar{\mu}_{n}^{*}}$ is the true canonical position, shared by
both sources since it encodes the agent's intrinsic geometry independent of
capture time. Setting $\vo{\bar{\mu}_{n}}=\vo{\bar{\mu}_{n}^{*}}$ therefore
yields $\delta_{n}^{c}=\mathbf{0}$ for \emph{both} sources simultaneously:
\begin{equation}
\label{eq:dst_zero}
  \vo{\bar{\mu}_{n}}=\vo{\bar{\mu}_{n}^{*}}
  \;\Longrightarrow\;
  \delta_{n}^{v}=\delta_{n}^{f}=\mathbf{0}
  \;\Longrightarrow\;
  \mathcal{L}_{\mathrm{DST}}^{*}
  =\sum_{n=1}^{N}\bigl(\mathcal{L}_{n}^{v}+\mathcal{L}_{n}^{f}\bigr)
  =0.
\end{equation}
The key mechanism is that the source-specific poses $\vb{P^{v}},\vb{P^{f}}$
absorb the temporal discrepancy: each pose places the agent at the correct
world position for its own capture time, so the canonical parameters need
only encode the agent's intrinsic shape. This is in direct contrast to the
single-timeline case, where no canonical parameter can simultaneously
satisfy $\delta_{n}^{v}=\mathbf{0}$ and $\delta_{n}^{f}=\mathbf{0}$
(Part~(ii)).

To analyze the optimization dynamics, we employ the Empirical Neural
Tangent Kernel (NTK) adapted to the Gaussian splatting renderer. Let
$\theta\in\mathbb{R}^{d}$ be the full optimizable parameter vector; the
empirical NTK is the $M\times M$ matrix with entries
\begin{equation}
\label{eq:ntk_def}
  \vb{\Theta}(\mathbf{x}_{k},\mathbf{x}_{l})
  =\bigl\langle
    \nabla_{\!\theta}\hat{I}^{c_{k}}(u_{k};\theta),\;
    \nabla_{\!\theta}\hat{I}^{c_{l}}(u_{l};\theta)
  \bigr\rangle,
\end{equation}
where $\mathbf{x}_{k}=(u_{k},c_{k})$ indexes pixel $u_{k}$ from source
$c_{k}$. Under gradient flow on the squared loss, the residual vector
$\mathbf{r}(t)=\hat{\mathbf{I}}(t)-\mathbf{I}^{*}$ evolves as
$\dot{\mathbf{r}}(t)\approx-\vb{\Theta}(0)\,\mathbf{r}(t)$ in the
linearized regime (Assumption~\ref{assump:linear}), so the spectral
properties of $\vb{\Theta}$ govern the optimization dynamics. Partitioning
the parameter vector as $\theta=(\vo{\bar{\bm{\mu}}},\vb{\bm{\xi}})$, where
$\vo{\bar{\bm{\mu}}}\in\mathbb{R}^{3N}$ collects the canonical means and
$\vb{\bm{\xi}}$ collects the pose parameters, the NTK decomposes additively:
$\vb{\Theta}
=\vb{\Theta}^{\vo{\bar{\bm{\mu}}}}+\vb{\Theta}^{\vb{\bm{\xi}}}$.

In the single-timeline formulation, the shared pose $\vb{\xi}\in\mathbb{R}^{6}$
couples both sources, giving the pose Jacobian
$J_{\vb{\xi}}^{\mathrm{single}}
=(\nabla_{\!\vb{\xi}}\hat{\mathbf{I}}^{v};\,
  \nabla_{\!\vb{\xi}}\hat{\mathbf{I}}^{f})
\in\mathbb{R}^{M\times 6}$
and the pose kernel
\begin{align}
\label{eq:ntk_pose_single}
  \vb{\Theta}_{\mathrm{single}}^{\vb{\xi}}
  =J_{\vb{\xi}}^{\mathrm{single}}\,(J_{\vb{\xi}}^{\mathrm{single}})^{\!\top}
  =\begin{pmatrix}
    \nabla_{\!\vb{\xi}}\hat{\mathbf{I}}^{v}
    (\nabla_{\!\vb{\xi}}\hat{\mathbf{I}}^{v})^{\!\top}
    & \nabla_{\!\vb{\xi}}\hat{\mathbf{I}}^{v}
      (\nabla_{\!\vb{\xi}}\hat{\mathbf{I}}^{f})^{\!\top}
    \\[6pt]
    \nabla_{\!\vb{\xi}}\hat{\mathbf{I}}^{f}
    (\nabla_{\!\vb{\xi}}\hat{\mathbf{I}}^{v})^{\!\top}
    & \nabla_{\!\vb{\xi}}\hat{\mathbf{I}}^{f}
      (\nabla_{\!\vb{\xi}}\hat{\mathbf{I}}^{f})^{\!\top}
  \end{pmatrix}.
\end{align}
The off-diagonal blocks
$\nabla_{\!\vb{\xi}}\hat{\mathbf{I}}^{v}
(\nabla_{\!\vb{\xi}}\hat{\mathbf{I}}^{f})^{\!\top}$
are generically non-zero: a pose update driven by the vehicle loss
simultaneously perturbs the infrastructure residuals, and vice versa.
This is the NTK-level manifestation of the gradient conflict identified
in Part~(ii).

Under DUST-GSG, the poses are parameterized by separate
$\vb{\xi^{v}}\in\mathbb{R}^{6}$ and $\vb{\xi^{f}}\in\mathbb{R}^{6}$.
Since the vehicle renderer uses only $\vb{P^{v}(\xi^{v})}$ and the
infrastructure renderer uses only $\vb{P^{f}(\xi^{f})}$, there is no
functional dependence between them:
$\partial\hat{I}^{v}(u)/\partial\vb{\xi^{f}}=\mathbf{0}$ and
$\partial\hat{I}^{f}(u)/\partial\vb{\xi^{v}}=\mathbf{0}$ for all $u$.
The pose Jacobian therefore has a block-diagonal structure:
\begin{equation}
\label{eq:jac_dst}
  J_{\vb{\xi}}^{\mathrm{DST}}
  =\begin{pmatrix}
    \nabla_{\!\vb{\xi^{v}}}\hat{\mathbf{I}}^{v} & \mathbf{0}\\[2pt]
    \mathbf{0} & \nabla_{\!\vb{\xi^{f}}}\hat{\mathbf{I}}^{f}
  \end{pmatrix}
  \in\mathbb{R}^{M\times 12},
\end{equation}
and the resulting pose kernel is
\begin{align}
\label{eq:thm_ntk}
  \vb{\Theta_{\mathrm{DST}}^{\bm{\xi}}}
  &= J_{\vb{\xi}}^{\mathrm{DST}}
     \,\bigl(J_{\vb{\xi}}^{\mathrm{DST}}\bigr)^{\!\top}
  \nonumber\\
  &= \begin{pmatrix}
       \nabla_{\!\vb{\xi^{v}}}\hat{\mathbf{I}}^{v}
       \bigl(\nabla_{\!\vb{\xi^{v}}}\hat{\mathbf{I}}^{v}\bigr)^{\!\top}
       & \mathbf{0}
       \\[6pt]
       \mathbf{0}
       & \nabla_{\!\vb{\xi^{f}}}\hat{\mathbf{I}}^{f}
         \bigl(\nabla_{\!\vb{\xi^{f}}}\hat{\mathbf{I}}^{f}\bigr)^{\!\top}
     \end{pmatrix}
  \nonumber\\
  &= \mathrm{diag}\!\left(
       \vb{\Theta_{vv}^{\xi^{v}}},\;
       \vb{\Theta_{ff}^{\xi^{f}}}
     \right),
\end{align}

where $\vb{\Theta_{vv}^{\xi^{v}}}
\coloneqq\nabla_{\!\vb{\xi^{v}}}\hat{\mathbf{I}}^{v}
(\nabla_{\!\vb{\xi^{v}}}\hat{\mathbf{I}}^{v})^{\!\top}
\in\mathbb{R}^{M_{v}\times M_{v}}$ and
$\vb{\Theta_{ff}^{\xi^{f}}}
\coloneqq\nabla_{\!\vb{\xi^{f}}}\hat{\mathbf{I}}^{f}
(\nabla_{\!\vb{\xi^{f}}}\hat{\mathbf{I}}^{f})^{\!\top}
\in\mathbb{R}^{M_{f}\times M_{f}}$.
The off-diagonal blocks vanish identically, establishing the
block-diagonal structure asserted in Eq.~\eqref{eq:thm_ntk}.

Partitioning the full residual vector as
$\mathbf{r}(t)=(\mathbf{r}^{v}(t)^{\!\top},\mathbf{r}^{f}(t)^{\!\top})^{\!\top}$,
the gradient flow restricted to pose parameters evolves as
\begin{align}
\label{eq:gf_full}
  \frac{\mathrm{d}}{\mathrm{d}t}
  \begin{pmatrix}\mathbf{r}^{v}\\\mathbf{r}^{f}\end{pmatrix}
  =
  -\vb{\Theta_{\mathrm{DST}}^{\bm{\xi}}}
  \begin{pmatrix}\mathbf{r}^{v}\\\mathbf{r}^{f}\end{pmatrix}
  =
  -\begin{pmatrix}
    \vb{\Theta_{vv}^{\xi^{v}}} & \mathbf{0}\\[4pt]
    \mathbf{0} & \vb{\Theta_{ff}^{\xi^{f}}}
  \end{pmatrix}
  \begin{pmatrix}\mathbf{r}^{v}\\\mathbf{r}^{f}\end{pmatrix}.
\end{align}
The block-diagonal structure decouples the two subsystems completely:
\begin{equation}
\label{eq:gf_decoupled}
  \dot{\mathbf{r}}^{v}(t)
  =
  -\vb{\Theta_{vv}^{\xi^{v}}}\,\mathbf{r}^{v}(t),
  \qquad
  \dot{\mathbf{r}}^{f}(t)
  =
  -\vb{\Theta_{ff}^{\xi^{f}}}\,\mathbf{r}^{f}(t).
\end{equation}
These two ODEs are independent: the closed-form solution
$\mathbf{r}^{v}(t)=e^{-\vb{\Theta_{vv}^{\xi^{v}}}t}\mathbf{r}^{v}(0)$
depends only on $\vb{\xi^{v}}$ and $\mathbf{r}^{v}(0)$, with no dependence
on $\vb{\xi^{f}}$ or $\mathbf{r}^{f}$, and vice versa. Consequently, a
gradient step on $\vb{\xi^{v}}$ driven by $\mathbf{r}^{v}$ produces
\emph{exactly zero} change in $\mathbf{r}^{f}$, and vice versa---the
precise sense in which DUST-GSG eliminates cross-source gradient
interference at the pose level. Combined with the zero-residual result
of Eq.~\eqref{eq:dst_zero}, this establishes all three claims of
Part~(iii). \hfill$\square$

\section{Implementation Details}
\label{app:implementation_details}

\subsection{Initialization}

For the background model, we combine $8 \times 10^5$ LiDAR points with $2 \times 10^5$ random samples, divided into $1 \times 10^5$ near samples uniformly distributed by distance to the scene's origin and $1
  \times 10^5$ far samples uniformly distributed by inverse distance. We filter out LiDAR samples belonging to dynamic objects. For rigid nodes and non-rigid deformable nodes, we utilize their tracking bounding boxes to accumulate LiDAR points. Each dynamic instance is initialized with at most $5\,000$ canonical Gaussians, and we only model moving instances, with trajectory length thresholds of $1.0$ for rigid nodes and $0.5$ for deformable nodes. To determine the
  initial color, LiDAR points are projected onto the image plane, whereas random samples are initialized with random colors.

  Prior to training, we perform offline Pose Correction to initialize the dual-timeline poses. Co-visible static vehicles serve as geometric anchors, matched across the vehicle and infrastructure views Euclidean distance and the Hungarian method. A 6-DoF pose correction is estimated by minimizing corner alignment error using L-BFGS. The refined poses regenerate cooperative labels, and missing tracks are filled via linear interpo
  lation for translation and Slerp for rotation for gaps up to two frames. These labels initialize separate pose trajectories for the vehicle-side and infrastructure-side timelines, with all timestamps normalized to $[0, 1]$.

\subsection{Training}

Our method trains for $30\,000$ iterations with all scene nodes optimized jointly. The learning rate for background Gaussian properties aligns with the default settings of 3DGS: positions at $1.6 \times 10^{-4}$ (d
  ecaying to $1.6 \times 10^{-6}$), SH DC at $2.5 \times 10^{-3}$, SH rest at $1.25 \times 10^{-4}$, opacity at $5 \times 10^{-2}$, scaling at $5 \times 10^{-3}$, and rotation at $1 \times 10^{-3}$. The degrees of spherical harmonics a
  reset to $3$ for background nodes, rigid nodes, and deformable nodes. For deformable nodes, the canonical position learning rate is scaled by a factor of $2.0$.

  The learning rate for the rotation of instance poses is $1 \times 10^{-5}$, decreasing exponentially to $5 \times 10^{-6}$. The learning rate for the translation of instance poses is $5 \times 10^{-4}$, decreasing exponentially to $1
  \times 10^{-4}$. For deformable nodes, the deformation network (8 layers, 256 channels) uses a learning rate of $1.6 \times 10^{-3}$, decaying to $1.6 \times 10^{-4}$, and the temporal embedding uses $1 \times 10^{-3}$, decaying to $
  1 \times 10^{-4}$. The sky environment map is optimized at $0.01$. Affine transformation and camera pose refinement modules both use $1 \times 10^{-5}$.

  For the Gaussian densification strategy, we utilize the absolute gradient of Gaussians. We set the densification threshold of the position gradient to $5 \times 10^{-4}$ and the densification size threshold to $3 \times 10^{-3}$. We
  reset the opacity of Gaussians to $0.01$ every $3\,000$ iterations, with an opacity warm-up of $500$ steps and refinement every $100$ steps. We stop splitting background Gaussians at $15\,000$ steps, rigid nodes at $30\,000$ steps, a
  nd deformable nodes at $20\,000$ steps.

  The photometric loss combines L1 and SSIM with weights $0.8$ and $0.2$. Additional regularization includes mask ($0.05$), depth ($0.01$), affine ($1 \times 10^{-5}$), and dynamic region loss ($0.05$, with $0.1$ for vehicles and $0.05
  $ for humans, activated from iteration $3\,000$). For pose regularization, temporal smoothness uses $0.01$ over $5$ frames; effective-translational temporal smoothness uses $0.002$ over $1$ frame; time-offset velocity loss is $0.01$;
  filled residual loss is $0.02$; and group residual loss is $0.01$ (stopped after $12\,000$ steps). The drift loss weight linearly decays to zero during training.

  The training process commences from a $1/4$ image resolution, doubling every $250$ iterations. We supervise with two asynchronous cameras (vehicle-side cam0 and roadside cam1), using independently provided timestamp files, with the roadside view as the reference for pose-group alignment. The sky environment map resolution is $1024$.

  Our method runs on NVIDIA A100 Tensor Core GPU, with training for each scene taking about one hour. Training time varies with different training settings.

\subsection{Optimization}
Since our newly constructed regularization term for pose optimization has already been introduced in Sec~\ref{sec:3_3}, only the remaining regularization terms are detailed here. Specifically, following the implementation of OMNIRE~\cite{chen2025omnire}, they are defined as follows:

We utilize the loss function introduced in Eq~\ref{eq:image_loss} to jointly optimize all learnable parameters. The image loss is computed as:
\begin{equation}
\label{eq:image_loss}
    \mathcal{L}_{\text{image}} = (1 - \lambda_r) \mathcal{L}_1 + \lambda_r \mathcal{L}_{\text{SSIM}}
\end{equation}
due to sparse temporal-spatial observation of the dynamic part, its supervision signal is insufficient. To address this, we apply a higher image loss weight to the dynamic regions identified by the rendered dynamic mask. This weight is set to 5. The depth map loss is computed as:
\begin{equation}
    \mathcal{L}_{\text{depth}} = \frac{1}{hw} \sum \left\| \mathcal{D}^s - \hat{\mathcal{D}} \right\|_1
\end{equation}
where $\mathcal{D}^s$ is the inverse of the sparse depth map. We project LiDAR points onto the image plane to generate the sparse LiDAR map, and $\hat{\mathcal{D}}$ is the inverse of the predicted depth map.

The mask loss $\mathcal{L}_{\text{opacity}}$ is computed as:
\begin{equation}
    \mathcal{L}_{\text{opacity}} = -\frac{1}{hw} \sum O_g \cdot \log O_g - \frac{1}{hw} \sum M_{\text{sky}} \cdot \log(1 - O_g)
\end{equation}
where $M_{\text{sky}}$ is the sky mask, and $O_g$ is the rendered opacity map.

In addition to the reconstruction losses, we introduce various regularization terms for different Gaussian representations to improve quality. Among these, an important regularization term is $\mathcal{L}_{\text{pose}}$, designed to ensure smooth human body poses $\boldsymbol{\theta}(t)$. This term is defined as:
\begin{equation}
    \mathcal{L}_{\text{reg}} = \frac{1}{2} \left\| \boldsymbol{\theta}(t - \delta) + \boldsymbol{\theta}(t + \delta) - 2\boldsymbol{\theta}(t) \right\|_1
\end{equation}
where $\delta$ is a randomly chosen integer from $\{1, 2, 3, 4, 5\}$. We set the weight of the SSIM loss, $\lambda_r$, to 0.2, the depth loss, $\lambda_{\text{depth}}$, to 0.01, the opacity loss, $\lambda_{\text{opacity}}$, to 0.05, and the pose smoothness loss, $\lambda_{\text{reg}}$, to 0.01.

\section{Evaluation}
\label{app:eva}

\subsection{Datasets}
\label{app:datasets}
We evaluate our method on 26 diverse sequences selected from the large-scale V2X-Seq dataset. These sequences cover various times of data, weather and traffic densities. Specifically, for the "times of data" category, night-time V2I scenarios are not provided in the V2X-Seq~\cite{v2x-seq} dataset. Thus, the period from 11:30 to 12:30 is defined as "Noon". The period from 16:30 to 17:30 is defined as "Dusk". Four sequences are selected for each time period. For the "Weather" category, only normal and rainy conditions are available in the dataset. Therefore, two subcategories are defined: "Normal" and "Rainy". Six sequences are allocated to each subcategory. For the "traffic densities" category, referred to as "Crowd", the number of agents in each sequence was counted. A higher number of agents indicates higher traffic densities. The 6 sequences with the most agents are selected as "High". The 6 sequences with the fewest agents are selected as "Low". The selected sequence id, average frames, average agent counts, and data formats for each category are detailed in Tables~\ref{tab:my_table4}. Examples for each category are illustrated in Fig~\ref{fig:appendix_fig1}.
\begin{table}[h]
  \centering
  \caption{Detailed descriptions of the sequences selected for classification are provided. The specific condition of the sequences is denoted by Cond. The average number of image frames across all sequences under each condition is represented by Frames. The average number of agents in all sequences under each condition is indicated by Agent.}
  \label{tab:my_table4}
  \begin{tabular}{llccl}
    \toprule
    Cond. & Sequence ID & Frames & Agent & Data Format \\
    \midrule
    Noon & 70,71,72,73 & 181 & 27 & Image, LiDAR, Bounding Box \\
    Dusk & 55,56,57,58 & 204 & 22 & Image, LiDAR, Bounding Box \\
    Normal & 21,23,29,30,60,63 & 176 & 17 & Image, LiDAR, Bounding Box \\
    Rainy & 36,40,41,42,47,48 & 166 & 10 & Image, LiDAR, Bounding Box \\
    Low & 7,21,23,36,42,79 & 156 & 5 & Image, LiDAR, Bounding Box \\
    High & 2,4,59,66,71,73 & 240 & 48 & Image, LiDAR, Bounding Box \\
    \bottomrule
  \end{tabular}
\end{table}

\begin{figure}
\centering
\includegraphics[width=1\linewidth]{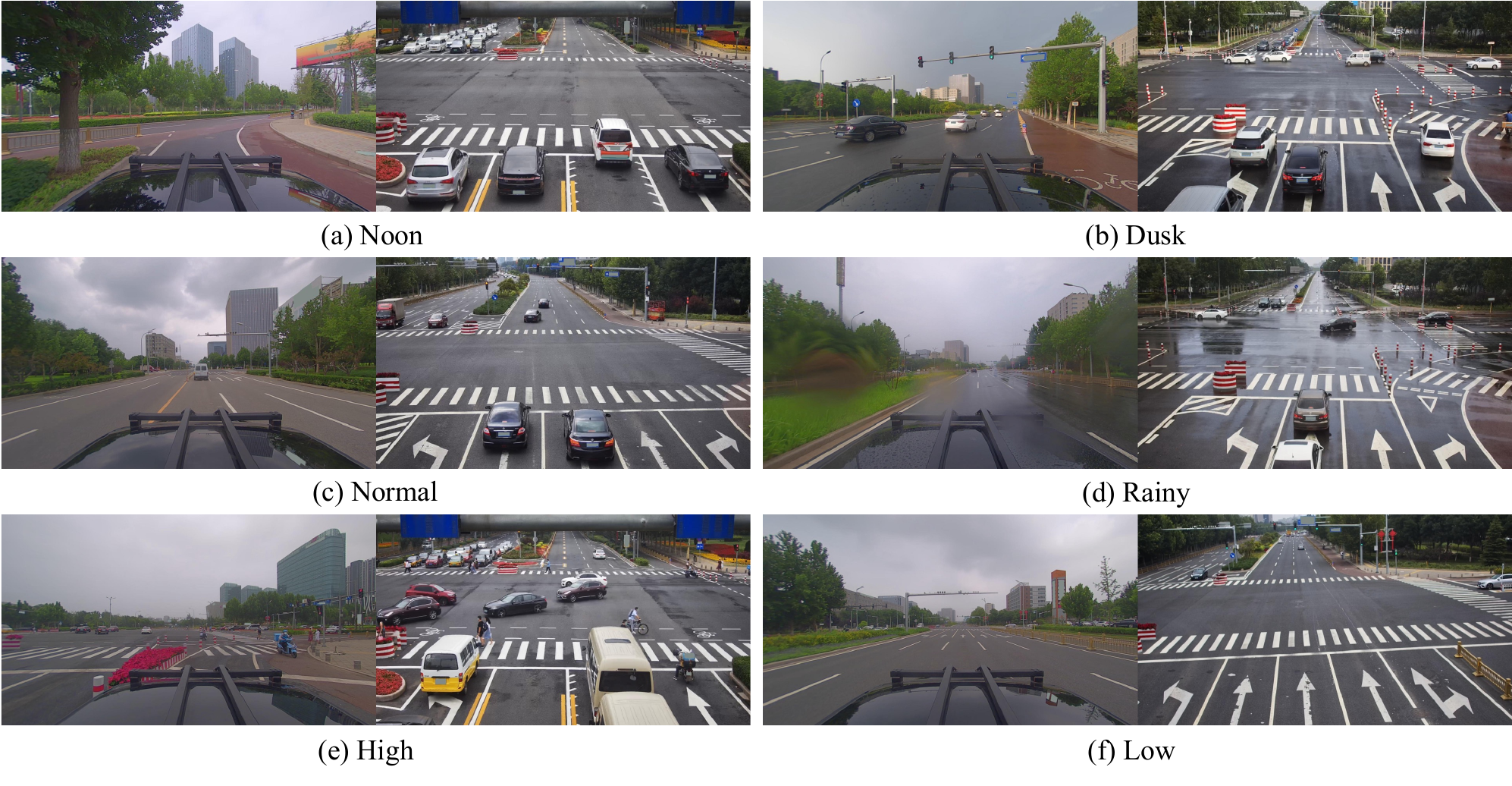}
\caption{The images from both the vehicle and infrastructure perspectives of the dataset are presented under each condition.}
\label{fig:appendix_fig1}
\end{figure}

\subsection{Metrics}
\label{appendix_metric}
We assess reconstruction quality using five metrics: PSNR~\cite{1284395}, SSIM, LPIPS~\cite{zhang2018perceptual}, FVD~\cite{unterthiner2018towards}, and RAFT-EPE~\cite{teed2020raft}.
  For the image-based metrics (PSNR, SSIM, and LPIPS), we report results on the full image as well as within dynamic regions. For temporal consistency, FVD and RAFT-EPE are computed exclusively within dynamic masks to avoid static back
  grounds dominating the motion-related evaluation.

  \vspace{0.5em}
  \noindent\textbf{PSNR.} The Peak Signal-to-Noise Ratio measures pixel-wise reconstruction fidelity between the rendered image $\hat{\mathbf{I}}$ and the ground-truth image $\mathbf{I}$. It is defined as:
  \begin{equation}
  \text{PSNR} = -10 \log_{10}\!\left( \frac{1}{N}\sum_{i=1}^{N}\bigl(\hat{I}_i - I_i\bigr)^2 \right),
  \end{equation}
  where $N$ is the number of valid pixels and pixel intensities are normalized to $[0,1]$. A higher PSNR indicates lower reconstruction error.

  \noindent\textbf{SSIM.} The Structural Similarity Index Measure evaluates local structural consistency by comparing luminance, contrast, and structure within a sliding Gaussian window. For two image patches $\mathbf{x}$ and $\mathbf{
  y}$, it is formulated as:
  \begin{equation}
  \text{SSIM}(\mathbf{x}, \mathbf{y}) = \frac{(2\mu_x\mu_y + c_1)(2\sigma_{xy} + c_2)}{(\mu_x^2 + \mu_y^2 + c_1)(\sigma_x^2 + \sigma_y^2 + c_2)},
  \end{equation}
  where $\mu_x,\mu_y$ and $\sigma_x,\sigma_y$ are the local means and standard deviations, $\sigma_{xy}$ is the cross-covariance, and $c_1,c_2$ are small constants for numerical stability. We compute the full SSIM map using an $11\times 11$ Gaussian window and average the values over the valid pixel region.

  \noindent\textbf{LPIPS.} The Learned Perceptual Image Patch Similarity measures perceptual distance using deep network features. Given rendered and ground-truth images, we extract multi-scale features from a pre-trained AlexNet and c
  ompute a normalized $\ell_2$ distance:
  \begin{equation}
  \text{LPIPS}(\hat{\mathbf{I}}, \mathbf{I}) = \sum_{\ell} \frac{1}{H_\ell W_\ell}\sum_{h,w} \bigl\| \mathbf{w}_\ell \odot \bigl( \phi_\ell(\hat{\mathbf{I}})_{h,w} - \phi_\ell(\mathbf{I})_{h,w} \bigr) \bigr\|_2^2,
  \end{equation}
  where $\phi_\ell(\cdot)$ denotes the feature map from layer $\ell$ of the network, $\mathbf{w}_\ell$ are learned channel-wise weights, and $H_\ell,W_\ell$ are the spatial dimensions of the feature map. The input images are normalized
  from $[0,1]$ to $[-1,1]$ before feature extraction. Lower LPIPS indicates better perceptual similarity.

  \noindent\textbf{FVD.} The Fréchet Video Distance measures temporal consistency by computing the Fréchet distance between the feature distributions of rendered and ground-truth videos. We extract spatio-temporal features using a pre-
  trained I3D network (trained on Kinetics) that yields 400-dimensional embeddings. For a video with $T$ frames, we build non-overlapping clips of length $T_c=16$ frames. Each clip is pre-processed by resizing the shorter side to 224 p
  ixels, center-cropping to $224\times 224$, and normalizing to $[-1, 1]$. Let $(\boldsymbol{\mu}_{\text{gen}}, \boldsymbol{\Sigma}_{\text{gen}})$ and $(\boldsymbol{\mu}_{\text{real}}, \boldsymbol{\Sigma}_{\text{real}})$ denote the mea
  n and covariance of the generated and real clip features, respectively. The FVD is defined as:
  \begin{equation}
  \text{FVD} = \|\boldsymbol{\mu}_{\text{gen}} - \boldsymbol{\mu}_{\text{real}}\|_2^2 + \operatorname{Tr}\!\left( \boldsymbol{\Sigma}_{\text{gen}} + \boldsymbol{\Sigma}_{\text{real}} - 2\sqrt{\boldsymbol{\Sigma}_{\text{gen}}\boldsymbol
  {\Sigma}_{\text{real}}} \right),
  \end{equation}
  where $\operatorname{Tr}(\cdot)$ is the matrix trace and $\sqrt{\cdot}$ denotes the matrix square root. Following standard practice, FVD is computed only within dynamic regions so that temporal consistency of moving agents is emphasi
  zed.

  \noindent\textbf{RAFT-EPE.} To evaluate motion plausibility, we compute the End-Point Error (EPE) between optical flows estimated from consecutive rendered frames and the corresponding ground-truth frames. We use RAFT with the pre-tr
  ained weights on FlyingThings ($\text{raft-things.pth}$) and run inference with 20 update iterations. For consecutive frames $\mathbf{I}_t$ and $\mathbf{I}_{t+1}$, RAFT predicts a dense flow field $\mathbf{F}_t\in\mathbb{R}^{H\times
  W\times 2}$ with horizontal and vertical components $(u,v)$. The EPE is computed as:
  \begin{equation}
  \text{EPE} = \frac{1}{|\mathcal{M}|} \sum_{(h,w)\in\mathcal{M}} \sqrt{\bigl(\hat{u}_{h,w} - u_{h,w}\bigr)^2 + \bigl(\hat{v}_{h,w} - v_{h,w}\bigr)^2},
  \end{equation}
  where $\mathcal{M}$ denotes the union of dynamic masks from both frames, ensuring evaluation focuses strictly on moving regions.

\subsection{Baselines}
\label{appendix_baseline}
To evaluate the effectiveness of our approach, our method is compared with several state-of-the-art baselines.

\begin{itemize}[leftmargin=*,nosep]
    \item \textbf{3DGS~\cite{kerbl20233d}}: 3D Gaussian Splatting (3DGS) is utilized for real-time and high-quality novel-view synthesis. Continuous volumetric radiance fields are represented by 3D Gaussians. These Gaussians are initialized from sparse calibration points. Interleaved optimization and density control are performed to accurately represent the scene geometry. A fast visibility-aware rendering algorithm is developed to support anisotropic splatting and real-time rendering.
    
    \item \textbf{PVG~\cite{chen2026periodic}}: Periodic Vibration Gaussian (PVG) is introduced to model dynamic urban scenes. A unified representation model is presented to capture the synergistic interactions of static and dynamic elements. Periodic vibration-based temporal dynamics are incorporated into the 3D Gaussian splatting technique. To handle sparse training data, a temporal smoothing mechanism and a position-aware adaptive control strategy are proposed. High-quality reconstruction is achieved without manually labeled bounding boxes or optical flow estimation.
    
    \item \textbf{OMNIRE~\cite{chen2025omnire}}: OMNIRE is presented as a comprehensive system for the creation of high-fidelity digital twins from on-device logs. Diverse dynamic objects, including vehicles and pedestrians, are fully reconstructed. Scene graphs are built on 3DGS. Multiple Gaussian representations are constructed in canonical spaces to model various dynamic actors. Advanced simulations with human-participated scenarios are supported by this holistic reconstruction capability.
    
    \item \textbf{CRUISE~\cite{xu2025cruise}}: CRUISE is proposed as a comprehensive reconstruction and synthesis framework for Vehicle-to-everything (V2X) driving environments. Real-world scenes are accurately reconstructed using decomposed Gaussian Splatting. Dynamic traffic participants are decomposed into editable Gaussian representations. Images from both ego-vehicle and infrastructure views are rendered to facilitate large-scale V2X dataset augmentation.
    
    \item \textbf{StreetGS~\cite{yan2024street}}: Street Gaussians (StreetGS) is introduced to model dynamic urban streets. The dynamic urban scene is explicitly represented as a set of point clouds equipped with semantic logits and 3D Gaussians. Foreground vehicles and the background are modeled separately. Tracked poses and a 4D spherical harmonics model are optimized to capture the dynamics and appearance of object vehicles. Fast scene editing and rendering are achieved by this explicit representation.
    
    \item \textbf{DeformableGS~\cite{yang2024deformable}}: Deformable 3D Gaussians Splatting (DeformableGS) is proposed to model monocular dynamic scenes. 3D Gaussians are learned in a canonical space. A deformation field is applied to capture the intricate details of moving objects. An annealing smoothing training mechanism is introduced to mitigate the impact of inaccurate poses. Real-time rendering and high visual quality are achieved through a differential Gaussian rasterizer.
\end{itemize}

\end{document}